\documentclass{article}

\usepackage[final]{neurips_2024}

\usepackage[utf8]{inputenc} 
\usepackage[T1]{fontenc}    
\usepackage{hyperref}       
\usepackage{url}            
\usepackage{booktabs}       
\usepackage{amsfonts}       
\usepackage{nicefrac}       
\usepackage{microtype}      
\usepackage{xcolor}         
\usepackage{amssymb}
\usepackage{amsmath}
\usepackage{mathtools}
\usepackage{amsthm}
\usepackage{bm}

\setcitestyle{square,numbers}

\title{Subject-driven Text-to-Image Generation via Preference-based Reinforcement Learning}

\author{
  Yanting Miao\\
  Department of Computer Science \\
  University of Waterloo, Vector Institute \\
  \texttt{y43miao@uwaterloo.ca} \\
  \And
  William Loh \\
  Depart of Computer Science \\
  University of Waterloo, Vector Institute \\
  \texttt{wmloh@uwaterloo.ca} \\
  \And
  Suraj Kothawade \\
  Google \\
  \texttt{skothawade@google.com} \\
  \And
  Pascal Poupart \\
  Depart of Computer Science \\
  University of Waterloo, Vector Institute \\
  \texttt{ppoupart@uwaterloo.ca} \\
  \And
  Abdullah Rashwan \thanks{This project was completed during work at Google} \\
  Google \\
  \texttt{arashwan@google.com} \\
  \And
  Yeqing Li \\
  Google \\
  \texttt{yeqing@google.com}
}

\begin{document}

\maketitle

\begin{abstract}
    Text-to-image generative models have recently attracted considerable interest, enabling the synthesis of high-quality images from textual prompts. However, these models often lack the capability to generate specific subjects from given reference images or to synthesize novel renditions under varying conditions. Methods like DreamBooth and Subject-driven Text-to-Image (SuTI) have made significant progress in this area. Yet, both approaches primarily focus on enhancing similarity to reference images and require expensive setups, often overlooking the need for efficient training and avoiding overfitting to the reference images. In this work, we present the $\lambda$-Harmonic reward function, which provides a reliable reward signal and enables early stopping for faster training and effective regularization. By combining the Bradley-Terry preference model, the $\lambda$-Harmonic reward function also provides preference labels for subject-driven generation tasks. We propose Reward Preference Optimization (RPO), which offers a simpler setup (requiring only $3\%$ of the negative samples used by DreamBooth) and fewer gradient steps for fine-tuning. Unlike most existing methods, our approach does not require training a text encoder or optimizing text embeddings and achieves text-image alignment by fine-tuning only the U-Net component. Empirically, $\lambda$-Harmonic proves to be a reliable approach for model selection in subject-driven generation tasks. Based on preference labels and early stopping validation from the $\lambda$-Harmonic reward function, our algorithm achieves a state-of-the-art CLIP-I score of 0.833 and a CLIP-T score of 0.314 on DreamBench. Our Pytorch implementation is available at \url{https://github.com/andrew-miao/RPO}.
\end{abstract}

\begin{figure}
    \centering
    \includegraphics[width=\linewidth, ]{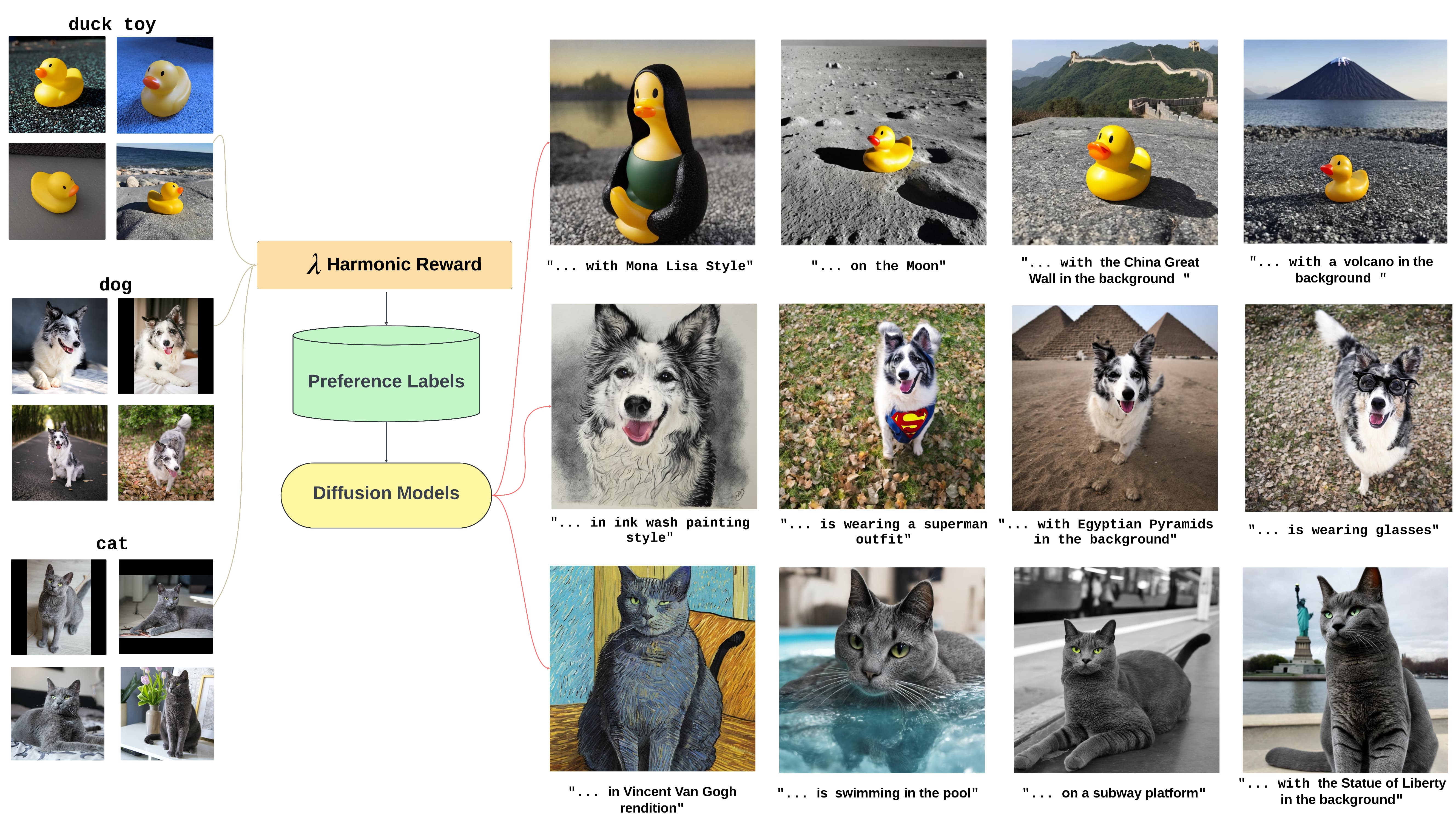}
    \caption{We illustrate the $\lambda$-Harmonic reward function applied to the subject-driven generation task. Leveraging preference labels produced by the $\lambda$-Harmonic reward function, alongside a few reference images, our preference-based algorithm efficiently generates unseen scenes that are both faithful to the reference images and the textual prompts.}
    \label{fig:show}
\end{figure}

\section{Introduction}

In the evolving field of generative AI, text-to-image diffusion models \cite{song2019generative, ho2020denoising, sohl2015deep, song2020score, song2021denoising, ruiz2023dreambooth, saharia2022photorealistic} have demonstrated remarkable abilities in rendering scenes that are both imaginative and contextually appropriate. However, these models often struggle with tasks that require the portrayal of specific subjects within text prompts. For instance, if provided with a photo of your cat, current diffusion models are unable to generate an image of your cat situated in the castle of your childhood dreams. This challenge necessitates a deep understanding of subject identity. Consequently, \textbf{\textit{subject-driven}} text-to-image generation has attracted considerable interest within the community. Chen et al. \cite{chen2024subject} have noted that this task requires complex transformations of reference images. Additionally, Ruiz et al. \cite{ruiz2023dreambooth} have highlighted that detailed and descriptive prompts about specific objects can lead to varied appearances in subjects. Thus, traditional image editing approaches and existing text-to-image models are ill-suited for subject-driven tasks.

Current subject-driven text-to-image generation methods are less expressive and expensive. Textual Inversion \cite{gal2022image} performs poorly due to the limited expressiveness of frozen diffusion models. Imagic \cite{kawar2023imagic} is both time-consuming and resource-intensive during the fine-tuning phase. It requires text-embedding optimization for each prompt, fine-tuning of diffusion models, and interpolation between optimized and target prompts. The training process is complex and slow. These text-based methods require 30 to 70 minutes to fine-tune their models, which is not scalable for real applications. SuTI \cite{chen2024subject} proposes an in-context learning method for subject-driven tasks. However, SuTI demands half a million expert models for each different subject, making it prohibitively expensive. Although SuTI can perform in-context learning during inference, the setup of expert models remains costly. DreamBooth \cite{ruiz2023dreambooth} provides a simpler method for handling subject-driven tasks. Nevertheless, DreamBooth requires approximately 1000 negative samples and 1000 gradient steps, and also needs fine-tuning of the text encoder to achieve state-of-the-art performance. Therefore, it is worthwhile to explore more efficient training methods: the setup should be as simple as possible. 
 First, training should not include multiple optimization phases. Second, text-to-image alignment should fine-tune as few components as possible such as the UNet, but not the text encoder for each prompt. Third, faster evaluation and regularization should be enabled by early stopping based on model selection.

In this paper, we propose a $\lambda$-Harmonic reward function that enables early stopping and accelerates training. In addition, we incorporate the Bradley-Terry preference model to generate preference labels. 
 We utilize preference-based reinforcement learning algorithms to finetune pre-trained diffusion models and to achieve text-to-image alignment without optimizing any text encoder or text embedding. The whole finetuning process including setup, training, validation, and model saving only takes 5 to 20 minutes on Cloud TPU V4. Our method, Reward Preference Optimization (RPO), only requires a few input reference images and the finetuned diffusion model can generate images that preserve the identity of a specific subject while aligning well with textual prompts (Figure \ref{fig:show}). 

To show the effectiveness of our $\lambda$-Harmonic reward function, we evaluate RPO on diverse subjects and text prompts on DreamBench \cite{ruiz2023dreambooth} and we report the DINO and CLIP-I/CLIP-T of RPO's generated images on this benchmark and compare them with existing methods. Surprisingly, our method requires a simple setup ($3\%$ of DreamBooth configuration) and with fewer gradient steps, but the experimental results outperform or match SOTA. 

In summary, our contributions are as follows:
\begin{itemize}
    \item We introduce the $\lambda$-Harmonic reward function, which permits early-stopping to alleviate overfitting in subject-driven generation tasks and to accelerate the finetuning process.
    \item By combining the $\lambda$-Harmonic reward function and a preference model, we present RPO, which only requires a cheap setup, but still can provide high quality results.
    \item We evaluate RPO and show the effectiveness of the $\lambda$-Harmonic function with diverse subjects and various prompts on DreamBench.  We achieve results comparable to SOTA.
\end{itemize}

\section{Related Works}

Ruiz et al. \cite{ruiz2023dreambooth} formulated a class of problems called \textit{subject-driven generation}, which refers to preserving the appearance of a subject contextualized in different settings. DreamBooth \cite{ruiz2023dreambooth} solves the issue of preserving the subject by binding it in textual space with a unique identifier for the subject in the reference images, and simultaneously generating diverse backgrounds by leveraging prior class-specific information previously learned. A related work that could possibly perform the same task is textual inversion \cite{gal2022image}. However, its original objective is to produce a modification of the subject or property marked by a unique token in the text. While it can be used to preserve the subject and change the background or setting, the performance is underwhelming compared to DreamBooth in various metrics \cite{ruiz2023dreambooth}.

The prevalent issue in DreamBooth and textual inversion is the long training time \cite{ruiz2023dreambooth, gal2022image} since gradient-based optimization has to be performed on their respective models for each subject. Subject-driven text-to-image generator (SuTI) by \cite{chen2024subject} aims to alleviate this issue by employing apprenticeship learning. By scraping millions of images online, many expert models are trained for different clusters of images centered around different subjects, which allows an apprentice to learn quickly from the experts \cite{chen2024subject}. However, this is an incredibly resource intensive task with massive computational overhead during training.

In the field of natural language processing, direct preference optimization has found great success in large language models (LLM) \cite{rafailov2024direct}. By bypassing reinforcement learning from human feedback and directly maximizing likelihoods using preference data, LLMs benefit from more stable training and reduced dependency on an external reward model. Subsequently, this inspired Diffusion-DPO by \cite{wallace2023diffusion}, which applies a similar technique to diffusion models. However,  this relies on a preference labelled dataset, which can be expensive to collect or not publicly available for legal reasons. 

Fortunately, there are reward models that can serve as functional substitutes such as CLIP \cite{radford2021learning} and ALIGN \cite{jia2021scaling}. ALIGN has a dual encoder architecture that was trained on a large dataset. The encoders can produce text and image embeddings, which allows us to obtain pairwise similarity scores by computing cosine similarity. There are also diffusion modelling techniques that can leverage reward models. An example is denoising diffusion policy optimization (DDPO) by Black et al. \cite{black2023training} that uses a policy gradient reinforcement learning method to encourage generations that lead to higher rewards.

\section{Preliminary}

In this section, we introduce the notation and some key concepts about text-to-image diffusion models and reinforcement learning.

\paragraph{Text-to-Image Diffusion Models.}

Diffusion models \cite{ho2020denoising, sohl2015deep, song2019generative, song2020score, song2021denoising} are a family of latent variable models of the form $\ptheta(\xv_0) = \int_\Xc \ptheta(\xv_{0:T}) d\xv_{1:T}$, where the $\xv_1, \dots, \xv_T$ are noised latent variables of the same dimensionality as the input data $\xv_0 \sim q(\xv_0)$. The diffusion or forward process is often a Markov chain that gradually adds Gaussian noise to the input data and each intermediate sample $\xv_t$ can be written as
\begin{align}\label{eq:preliminary_diffusion}
    \xv_t = \sqrt{\alpha_t} \xv_0 + \sqrt{1 - \alpha_t} \epsilonv_t, \quad \text{for all } t \in \{1, \dots, T\},
\end{align}
where $\alpha_t$ refers to the variance schedule and $\epsilonv_t \sim \Nc(\mathbf{0}, \Iv)$. Given a conditioning tensor $\cv$ (often a text embedding), the core premise of text-to-image diffusion models is to use a neural network $\epsilonv_{\thetav}(\xv_t, \cv, t)$ that iteratively refines the current noised sample $\xv_t$ to obtain the previous step sample $\xv_{t - 1}$, This network can be trained by optimizing a simple denoising objective function, which is the time coefficient weighted mean squared error, the derivation is shown in Appendix \ref{section:appendix_background}:
\begin{align}\label{eq:preliminary_diffusion_training}
    \Eb_{\xv_0, \cv, t, \epsilonv_t} \left[\omega(t) \lVert \epsilonv_{\thetav}(\xv_t, \cv, t) - \epsilonv_t \rVert^2_2\right],
\end{align}
where $t$ is uniformly sampled from $\{1, \dots, T\}$ and $\omega(t)$ is a time dependent weight that can be simplified to 1 according to \cite{ho2020denoising, song2021denoising, rombach2022high}.

\paragraph{Reinforcement Learning and Diffusion DPO} Reinforcement Learning  for diffusion models \cite{black2023training, fan2023reinforcement, wallace2023diffusion} aims to solve the following optimization problem:
\begin{align} \label{eq:method_rl_obj}
    \Eb_{\xv_{0:T} \sim \ptheta(\xv_{0:T} \mid \cv)}\left[\sum_{t=1}^{T} R(\xv_t, \xv_{t-1}, \cv) - \beta \Db_{\text{KL}}(\ptheta(\xv_{t - 1} \mid \xv_t, \cv) \| \pbase(\xv_{t - 1}\mid \xv_t, \cv))\right],
\end{align}
where $\beta$ is a hyperparameter controlling the KL-divergence between the finetuned model $\ptheta$ and the pre-trained base model $\pbase$. In Equation (14) from Diffusion-DPO \cite{wallace2023diffusion}, the optimal $\ptheta$ can be approximated by minimizing the negative log-likelihood:
\begin{align}\label{eq:method_dpo_diffusion}
    \Eb_{\xv^+_0, \xv^-_0, t, \xv^+_t, \xv^-_t} &\Bigg[-\log \sigma \bigg(\beta \big(\lVert \epsilonv_{\text{base}}(\xv^+_t, \cv, t) - \epsilonv^+_t \rVert^2_2 - \lVert \epsilonv_{\thetav}(\xv^+_t, \cv , t) - \epsilonv^+ \rVert^2_2  \nonumber \\ 
    & - (\lVert \epsilonv_{\text{base}}(\xv^-_t, \cv, t) - \epsilonv^-_t \rVert^2_2 - \lVert \epsilonv_{\thetav}(\xv^-_t, \cv , t) - \epsilonv^- \rVert^2_2) \big)\bigg)\Bigg],
\end{align}

where $\epsilonv^+$ and $\epsilonv^-$ are independent samples from a Gaussian distribution, $\xv^+_t$ and $\xv^-_t$ are perturbed versions of $\xv^+_0$ and $\xv^-_0$ that depend on $\epsilonv^+$ and $\epsilonv^-$, and $\xv^+_0$ is preferred to $\xv^-_0$.  A detailed description is given in Appendix~\ref{section:appendix_background}.

\paragraph{Additional notation.} We use $\xref$ and $\xgen$ to represent the reference image and generated image, respectively. $\Iref$ denotes the set of reference images, and $\Igen$ is the set of generated images. $\Pb(\xv \succ \tilde{\xv})$ represents the probability that $\xv$ is more preferred than $\tilde{\xv}$.

\section{Method}
We present our $\lambda$-Harmonic reward function that provides reward signals for subject-driven tasks to reduce the risk that the learned model will overfit to the reference images. Based on this reward function, we use the Bradley-Terry model to sample preference labels and a preference algorithm to finetune the diffusion model by optimizing both a similarity loss and a preference loss.

\subsection{Reward Preference Optimization}

In contrast to other fine-tuning applications \cite{wallace2023diffusion, ouyang2022training, rafailov2024direct, rafailov2024r}, there is no human feedback in the subject-driven text-to-image generation task. The model only receives a few reference images and a prompt with a specific subject. Hence, we first propose the $\lambda$-Harmonic reward function that can leverage the ALIGN model \cite{jia2021scaling} to provide feedback based on the generated image fidelity: similarity to the given reference images and faithfulness to the text prompts. 

\paragraph{$\lambda$-Harmonic Reward Function.} The normalized ALIGN-I and ALIGN-T scores can be denoted as
\begin{align*}
    &\text{ALIGN-I}(\xv, \Iref) \coloneq \frac{1}{|\Iref|}\sum_{\tilde{\xv} \in \Iref}\frac{\text{CosSim}(f_{\phiv}(\xv), f_{\phiv}(\tilde{\xv})) + 1}{2}  & \mbox{ (Image alignment)} \\
    &\text{ALIGN-T}(\xv, \cv) \coloneq \frac{\text{CosSim}(f_{\phiv}(\xv), g_{\phiv}(\cv)) + 1}{2} & \mbox{(Text alignment)},
\end{align*}
where CosSim is the cosine similarity, $f_{\phiv}(\xv)$ is the image feature extractor and $g_{\phiv}(\cv)$ is the text encoder in the ALIGN model. Given a reference image set $\Iref$, the $\lambda$-Harmonic reward function can be defined by a weighted harmonic mean of the ALIGN-I and ALIGN-T scores,
\begin{align}\label{eq:method_reward_fn}
    r(\xv, \cv; \lambda, \Iref) \coloneq \dfrac{1}{\dfrac{\lambda}{\text{ALIGN-I}(\xv, \Iref)} + \dfrac{1 - \lambda}{\text{ALIGN-T}(\xv, \cv)}}.
\end{align}
Compared to the arithmetic mean, there are two advantages to using the harmonic mean: (1) according to AM-GM-HM inequalities \cite{djukic2006imo}, the harmonic mean is a lower bound of the arithmetic mean and maximizing this ``pessimistic'' reward can also improve the arithmetic mean of ALIGN-I and ALIGN-T scores; (2) the harmonic mean is more sensitive to the smaller of the two scores, i.e., a larger reward is only achieved when both scores are relatively large. 

For a simple example, consider $\lambda = 0.5$. If there are two images, $\xv$ and $\tilde{\xv}$, where the first image achieves an ALIGN-I score of 0.9 and an ALIGN-T score of 0.01, and the second image receives an ALIGN-I score of 0.7 and an ALIGN-T score of 0.21, we may prefer the second image because it has high similarity to the reference images and is faithful to the text prompts. However, using the arithmetic mean would assign both images the same reward of 0.455. In contrast, the harmonic mean would assign the first image a reward of 0.020 and the second image a reward of 0.323, aligning with our preferences. During training, we set $\lambda_{\text{train}} = 0$, which means the reward model will focus solely on text-to-image alignment because the objective function consists only of a loss for image-to-image alignment. Note that we set $\lambda_{\text{val}}$ to a different value for validation, which evaluates the fidelity of the subject and faithfulness of the prompt. Details can be found in Section \ref{section:experiments}.

\paragraph{Dataset.} The set of images for subject-driven generative tasks can usually be represented as $\Ic = \Iref \cup \Igen$, where $\Igen$ is the image set generated by the base model. DreamBooth \cite{ruiz2023dreambooth} requires two different prompts, $\cv$ and $\cpr$, where $\cv$ includes a reference to the subject while $\cpr$ refers to the prior class of the subject but not the subject. For example, $\cv$ can be \texttt{``a photo of [V] dog''} and $\cpr$ can be \texttt{``a photo of a dog''}, where \texttt{``[V]''} is a unique token that refers to the subject and dog is the prior class of the subject. DreamBooth then uses $\cpr$ to generate a variety of images $\Igen$ in the prior class to avoid overfitting to the reference images via a regularizer. Typically, the size of the set of generated images is around 1000, i.e., $\lvert \Igen \rvert = 1000$, which is time-consuming and space-intensive in real applications. However, the diffusion model can only maximize the similarity score and still receives a high reward based on this uninformative prompt $\cpr$. Our method aims to balance the trade-off between similarity and faithfulness. Thus, for efficiency, we introduce 8 novel training prompts, $\cv_{\text{mod}}$ of the form \texttt{``a [V] [class noun] [modification]''} where modification includes \textit{artistic style transfer, re-contextualization}, and \textit{accessorization}.  For example, $\cv_{\text{mod}}$ can be \texttt{``a [V] dog is on the Moon''}.  These training prompts can be pre-specified or generated by other Large Language Models\footnote{SuTI \cite{chen2024subject} utilizes PaLM \cite{chowdhery2023palm} to generate unseen prompts during training}. The full list of training prompts is provided in the supplementary material (Figure~\ref{fig:appendix_training_prompts}).  We feed these training prompts to the base model and generate 4 images for each training prompt, i.e., $\lvert \Igen \rvert = 32$.

Once we obtain reward signals, we adopt the Bradley-Terry model \cite{bradley1952rank} to generate preference labels. In particular, given a tuple $(\xref, \xgen, \cv_{\text{mod}})$, we sample preference labels $y$ from the following probability model:
\begin{align}\label{eq:method_preference_model}
    \Pb(\xref \succ \xgen) \coloneq \frac{\exp(r(\xref, \cv_{\text{mod}}; \lambda, \Iref))}{\exp(r(\xref, \cv_{\text{mod}}; \lambda, \Iref)) + \exp(r(\xgen, \cv_{\text{mod}}; \lambda, \Iref))}.
\end{align}

\paragraph{Learning.}
The learning objective function consists of two parts --- similarity loss and preference loss. The similarity loss is designed to minimize the KL divergence between the distribution of reference images and the learned distribution $\ptheta(\xv)$, which is equivalent to minimizing:
\begin{align}\label{eq:method_similar_loss}
    \Lc_{\text{sim}}(\thetav) \coloneq \Eb_{\xref, \cv, t, \epsilonv_{\text{ref}}} \left[\lVert \epsilonv_{\thetav}(\xv_{\text{ref}, t}, \cv, t) - \epsilonv_{\text{ref}}\rVert^2_2\right], \quad t \sim \Uc\{1, \dots, T\}, \quad  \epsilonv_{\text{ref}} \sim \Nc(\mathbf{0}, \Iv).
\end{align}

The preference loss aims to capture the preference signals and fit the preference model, Eq. (\ref{eq:method_preference_model}). Therefore, we use binary cross-entropy as the objective function for the preference loss. Combining the DPO objective function in Eq. \ref{eq:method_dpo_diffusion}, the loss function can be written as follows:
\begin{align}\label{eq:method_preference_loss}
    \Lc_{\text{pref}}(\thetav) & \coloneq \Eb_{\xref, \xgen, \cv_{\text{mod}}, y, t, \epsilonvref, \epsilonvgen} \Bigg[ y \log \sigma \bigg(\beta \ell_{\thetav}(\xref, \xgen, \cv_{\text{mod}}, y, t, \epsilonvref, \epsilonvgen)\bigg) \nonumber \\
    & + (1 - y) \log \sigma \bigg(- \beta \ell_{\thetav}(\xref, \xgen, \cv_{\text{mod}}, y, t, \epsilonvref, \epsilonvgen) \bigg)\Bigg],
\end{align}
where
\begin{align*}
    \ell_{\thetav}(\xref, \xgen, \ \cv_{\text{mod}}, y, t, \epsilonvref, \epsilonvgen) &\coloneq \lVert \epsilonvbase(\xreft, \cv_{\text{mod}}, t) - \epsilonvref \rVert^2_2 - \lVert \epsilonv_{\thetav}(\xreft, \cv_{\text{mod}}, t) - \epsilonvref \rVert^2_2 \\
    &-(\lVert \epsilonvbase(\xgent, \cv_{\text{mod}}, t) - \epsilonvgen \rVert^2_2 - \lVert \epsilonv_{\thetav}(\xgent, \cv_{\text{mod}}, t) - \epsilonvgen \rVert^2_2),
\end{align*}
and $t \sim \Uc\{1, \dots, T\}$ and $\epsilonvref, \epsilonvgen \sim \Nc(\mathbf{0}, \Iv).$ Combining these two loss functions together, the objective function for finetuning is written as
\begin{align}\label{eq:method_total_obj}
    \Lc(\thetav) = \Lc_{\text{sim}}(\thetav) + \Lc_{\text{pref}}(\thetav)
\end{align}
\begin{figure}
    \centering
    \includegraphics[width=\textwidth]{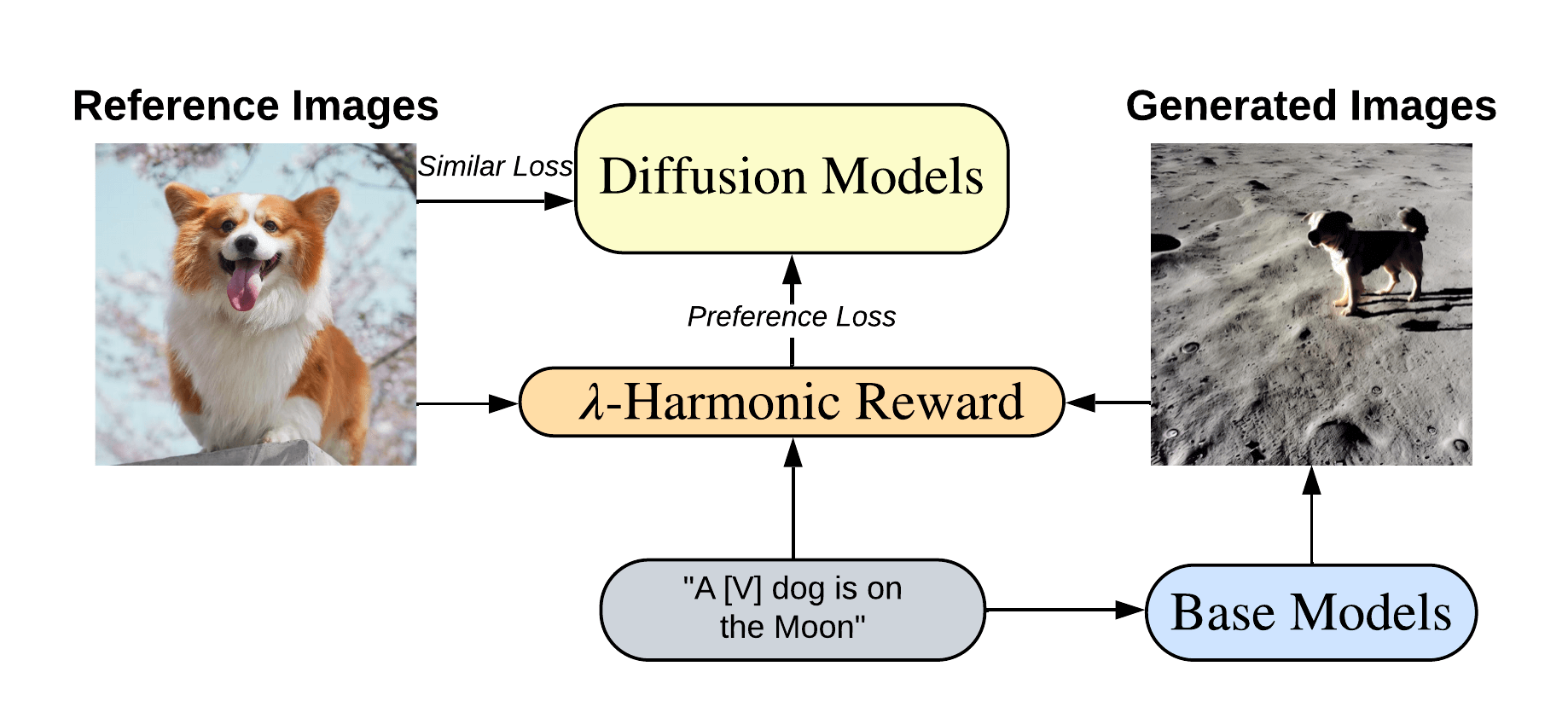}
    \caption{Overview of the finetuning phase for RPO. First, the base diffusion model generates a few images based on novel training prompts. Second, we compute the rewards for both reference and generated images using Equation (\ref{eq:method_reward_fn}). Then, preference labels are sampled according to the preference distribution, as defined in Equation (\ref{eq:method_preference_model}). Finally, the diffusion model is trained by minimizing both the similarity loss (Equation (\ref{eq:method_similar_loss})) and preference loss (Equation (\ref{eq:method_preference_loss})).}
    \label{fig:train_overview}
\end{figure}
Figure \ref{fig:train_overview} presents an overview of the training method, which includes the base model generated samples, the ALIGN reward model, and the preference loss. Note that $\Lc_{\text{pref}}$ serves as a regularizer for approximating the text-to-image alignment policy. Conversely, DreamBooth \cite{ruiz2023dreambooth} adopts $\Db_{\text{KL}}(p_{\text{base}}(\xv_{t - 1} \mid \xv_t, \cv) | \ptheta(\xv_{t - 1} \mid \xv_t, \cv))$ as its regularizer, which cannot guarantee faithfulness to the text-prompt. Based on this loss function and preference model, we only need a few hundred gradient steps and a small set size of $\lvert \Igen \rvert$ to achieve results that are comparable to, or even better than, the state of the art. The fine-tuning process, which includes generating images, training, and validation, takes about 5 to 20 minutes on a single Google Cloud Platform TPUv4-8 (32GB) for Stable Diffusion.

\section{Experiments}\label{section:experiments}

In this section, we present the experimental results demonstrated by RPO. We investigate several questions. First, can our algorithm learn to generate images that are faithful both to the preference images and to the textual prompts, according to preference labels? Second, if RPO can generate high-quality images, which part is the key component of RPO: the reference loss or the early stopping by the $\lambda$-Harmonic reward function? Third, how do different $\lambda_{\text{val}}$ values used during validation affect performance in RPO? We refer readers to Appendix \ref{section:appendix_experiments_details} for details on the experimental setup, Appendix \ref{section:appendix_skillset} for the skill set of RPO, Appendix \ref{section:appendix_limitations} for the limitations of the RPO algorithm, and Appendix \ref{section:appendix_future_work} for future work involving RPO.

\subsection{Dataset and Evaluation}

\paragraph{DreamBench.} In this work, we use the DreamBench dataset proposed by DreamBooth \citep{ruiz2023dreambooth}. This dataset contains 30 different subject images including backpacks, sneakers, boots, cats, dogs, and toy, etc. DreamBench also provides 25 various prompt templates for each subject and these prompts are requiring the learned models to have the following abilities: re-contextualization, accessorization, property modification, and attribute editing. 

\paragraph{Evaluation Metrics.} We follow DreamBooth \cite{ruiz2023dreambooth} and SuTI \cite{chen2024subject} to report DINO \cite{caron2021emerging}\footnote{DINO encodes only images and makes a better fine-grained understanding of images than CLIP. But it cannot provide signals on text-image alignment.} and CLIP-I \cite{radford2021learning} for evaluating image-to-image similarity score and CLIP-T \cite{radford2021learning} for evaluating the text-to-image similarity score. We also use our $\lambda$-Harmonic reward as a evaluation metric for the overall fidelity and the default value of $\lambda = 0.3$. For evaluation, we follow DreamBooth \cite{ruiz2023dreambooth} and SuTI \cite{chen2024subject} to generate 4 images per prompt, 3000 images in total, which provides a robust evaluation. 

\paragraph{Baseline algorithms.}  DreamBooth \cite{ruiz2023dreambooth}: A test-time fine-tuning method. This algorithm requires approximately $\lvert \Igen\rvert = 1000$ and 1000 gradient steps to finetune the UNet and text-encoder components. SuTI \cite{chen2024subject}: A pre-trained method that requires half a million expert models and introduces cross-attention layers into the original diffusion models. Textual Inversion \cite{gal2022image}: A text-based method that optimizes the text embedding but freezes the diffusion models. Re-Imagen \cite{chen2022re}: An information retrieval-based algorithm that modifies the backbone network architectures and introduces cross-attention layers into the original diffusion models. DisenBooth \cite{chen2023disenbooth}: A test-time fine-tuning method that generates subject-driven images by optimizing textual identity-preserving embeddings. Custom Diffusion \cite{kumari2023multi}: a test-time fine-tuning method, focused on efficient training by optimizing the key and value projection matrices in the cross-attention layers of diffusion models. ELITE \cite{wei2023elite}: A test-time fine-tuning approach consisting of two stages: one for training the textual embeddings and another for preserving identity. IP-Adapter \cite{ye2023ip}: a pretrained method where the decoupled cross-attention layer captures the textual signal while integrating reference images. SSR-Encoder \cite{zhang2024ssr}: also a pretrained method that highlights selective regions and extracts detailed features for subject-driven image generation.

\begin{figure}
    \centering
    \includegraphics[width=\textwidth]{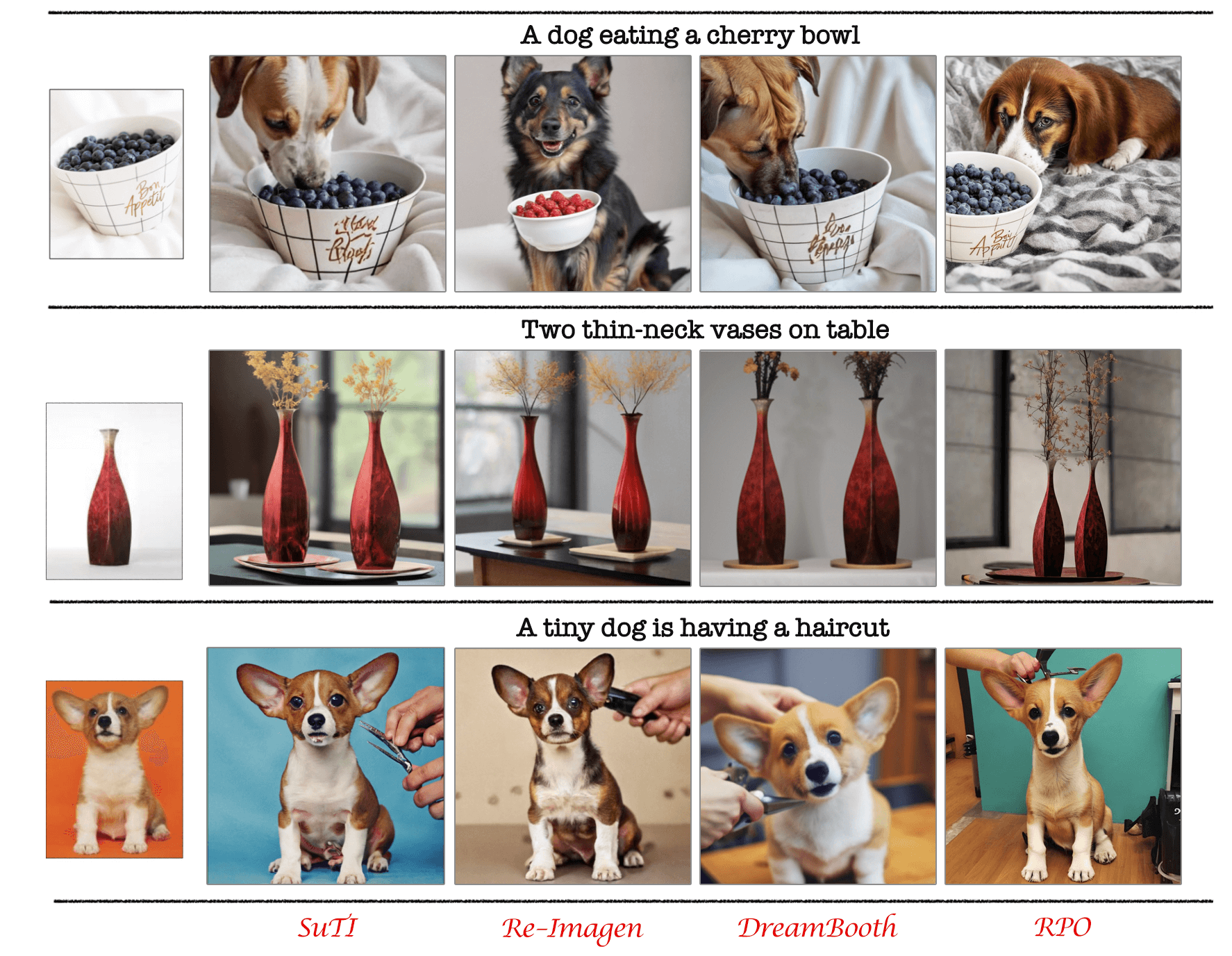}
    \caption{Qualitative comparison with other subject-driven text-to-image methods, adapted from \cite{chen2024subject}}
    \label{fig:comparison}
\end{figure}

\subsection{Results}
\paragraph{Quantitative Results.} We begin by addressing the first question. We use a quantitative evaluation to compare RPO with other existing methods on three metrics (DINO, CLIP-I, CLIP-T) in DreamBench to validate the effectiveness of RPO. The experimental results on DreamBench is shown in Table \ref{tab:dreambench_results}. We observe that RPO can perform better or on par with existing works on all three metrics. Compared to DreamBooth, RPO only requires $3\%$ of the negative samples, but RPO can outperform DreamBooth on the CLIP-I and CLIP-T scores by $3\%$ given the same backbone. Our method outperforms all baseline algorithms in the CLIP-T score, establishing a new SOTA result. This demonstrates that RPO, by solely optimizing UNet through preference labels from the $\lambda$-Harmonic reward function, can generate images that are faithful to the input prompts. Similarly, our CLIP-I score is also the highest, which indicates that RPO can generate images that preserve the subject's visual features. In terms of the DINO score, our method is almost the same as DreamBooth when using the same backbone. We conjecture that the reason RPO achieves higher CLIP scores and lower DINO score is that the $\lambda$-Harmonic reward function prefers to select images that are semantically similar to the textual prompt, which may result in the loss of some unique features in the pixel space.

\begin{table}
  \caption{Quantitative comparison for the number of iterations, subject fidelity and prompt fidelity.}
  \label{tab:dreambench_results}
  \centering
  \begin{tabular}{lll|lll}
    \toprule
    Method & Backbone &Iterations $\downarrow$ & DINO $\uparrow$ & CLIP-I $\uparrow$ & CLIP-T $\uparrow$ \\
    \midrule
    Reference Images & N/A &N/A & $0.774$ & $0.885$ & N/A \\
    \midrule
    DreamBooth \cite{ruiz2023dreambooth} & Imagen \cite{saharia2022photorealistic} & $1000$ & $0.696$ & $0.812$ & $0.306$ \\
    DreamBooth \cite{ruiz2023dreambooth} & SD \cite{rombach2022high} & $1000$ & $0.668$ & $0.803$ & $0.305$ \\
    Textual inversion \cite{gal2022image} & SD \cite{rombach2022high} & $5000$  & $0.569$ & $0.780$ & $0.255$\\
    SuTI \cite{chen2024subject} & Imagen \cite{saharia2022photorealistic} &$1.5 \times 10^5$ & $\bm{0.741}$ & $0.819$ & $0.304$ \\
    Re-Imagen \cite{chen2022re} & Imagen \cite{saharia2022photorealistic} & $2 \times 10^5$ & $0.600$ & $0.740$ & $0.270$ \\
    DisenBooth \cite{chen2023disenbooth} & SD\cite{rombach2022high} & $3000$ & $0.574$ & $0.755$ & $0.255$ \\
    Custom Diffusion \cite{kumari2023multi} & SD\cite{rombach2022high} & 500 & $0.695$ & $0.801$ & $0.245$ \\
    ELETE \cite{wei2023elite} & SD\cite{rombach2022high} & $3000$ & $0.652$ & $0.765$ & $0.255$ \\
    IP-Adapter \cite{ye2023ip} & SD\cite{rombach2022high} & $10^6$ & $0.608$ & $0.809$ & $0.274$ \\
    SSR-Encoder \cite{zhang2024ssr} & SD\cite{rombach2022high} & $10^6$ & $0.612$ & $0.821$ & $\bm{0.314}$ \\
    \midrule
    Ours: RPO & SD \cite{rombach2022high} & $\bm{400}$& $0.652$ & $\bm{0.833}$ & $\bm{0.314}$ \\
    \bottomrule
  \end{tabular}
\end{table}

\paragraph{Qualitative Results.}

We use the same prompt as SuTI \cite{chen2024subject}, and the generated images are shown in Figure \ref{fig:comparison}. RPO generates images that are faithful to both reference images and textual prompts. We noticed a semantic mistake in the first prompt used by SuTI \cite{chen2024subject}; it should be \texttt{A dog eating a cherry from a bowl}. Furthermore, each reference bowl image contains blueberries, and the ambiguous prompt caused the RPO-trained model to become confused during the inference phase. However, RPO still preserves the unique appearance of the bowl. For instance, while the text on the bowl is incorrect or blurred in the SuTI and DreamBooth results, RPO accurately retains the words \textit{Bon Appetit} from the reference bowl images. 
Although existing methods can produce images highly faithful to the reference images, they may not align as well with the textual prompts. We also provide an example in Appendix~\ref{section:appendix_addition_comparison} that shows how RPO can handle the failure case observed in DreamBooth and SuTI. Additionally, we find that the RPO model can generate reasonable images even for highly imaginative prompts, as shown in Appendix \ref{section:appendix_skillset} (Figure ~\ref{fig:novel_hybrids} and Figure ~\ref{fig:novel_prompts_dog}). These images demonstrate that the RPO-trained model does not overfit the training data. Instead, the $\lambda$-Harmonic function provides a method for selecting a model capable of text-to-image alignment, even when the textual prompts are highly imaginative.

\begin{table}
    \centering
    \caption{Ablation study on regularization to evaluate fidelity across multiple subjects and prompts. Standard deviation is included.}
    \begin{tabular}{lllll}
         \toprule
         Method & DINO $\uparrow$ & CLIP-I $\uparrow$ & CLIP-T $\uparrow$ & $0.3$-Harmonic $\uparrow$\\
         \midrule
         Pure $\Lc_{\text{sim}}$ & $\bm{0.695 \pm 0.077} $ & $\bm{0.852 \pm 0.043}$ & $0.285 \pm 0.027$ & $0.660 \pm 0.016$ \\
         $\Lc_{\text{pref}}$ w/o early-stopping & $0.688 \pm 0.082$ & $0.845 \pm 0.042$ & $0.296 \pm 0.027$ & $0.663 \pm 0.014$ \\
         Early-stopping w/o $\Lc_{\text{pref}}$ & $0.575 \pm 0.124$ & $0.799 \pm 0.052$ & $0.323 \pm 0.025$ & $0.672 \pm 0.013$\\
         RPO ($\lambda_{\text{val}} = 0.3$) & $0.581 \pm 0.113$ & $0.798 \pm 0.039$ & $\bm{0.329 \pm 0.021}$ & $\bm{0.673 \pm 0.013}$\\
         \bottomrule
    \end{tabular}
    \label{tab:ablation_study}
\end{table}

\begin{table}
  \caption{Different validation $\lambda_{\text{val}}$-Harmonic reward comparison for evaluating fidelity over multiple subjects and prompts. Standard deviation is included.}
  \label{tab:lambda_comparison}
  \centering
  \begin{tabular}{lllll}
    \toprule
    Configuration & DINO $\uparrow$ & CLIP-I $\uparrow$ & CLIP-T $\uparrow$ & $0.3$-Harmonic $\uparrow$\\
    \midrule
    $\lambda_{\text{val}} = 0.3$ & $0.581 \pm 0.113$ & $0.798 \pm 0.039$ & $\bm{0.329 \pm 0.021}$ & $\bm{0.673 \pm 0.013}$\\
    $\lambda_{\text{val}} = 0.5$ & $0.652 \pm 0.082$  & $0.833 \pm 0.041$ & $0.314 \pm 0.022$ & $0.671 \pm 0.008$ \\
    $\lambda_{\text{val}} = 0.7$ & $\bm{0.679 \pm 0.085}$ & $\bm{0.850 \pm 0.045}$ & $0.304 \pm 0.023$ & $0.667 \pm 0.011$ \\
    \bottomrule
  \end{tabular}
\end{table}

\begin{figure}
    \centering
    \includegraphics[width=\textwidth]{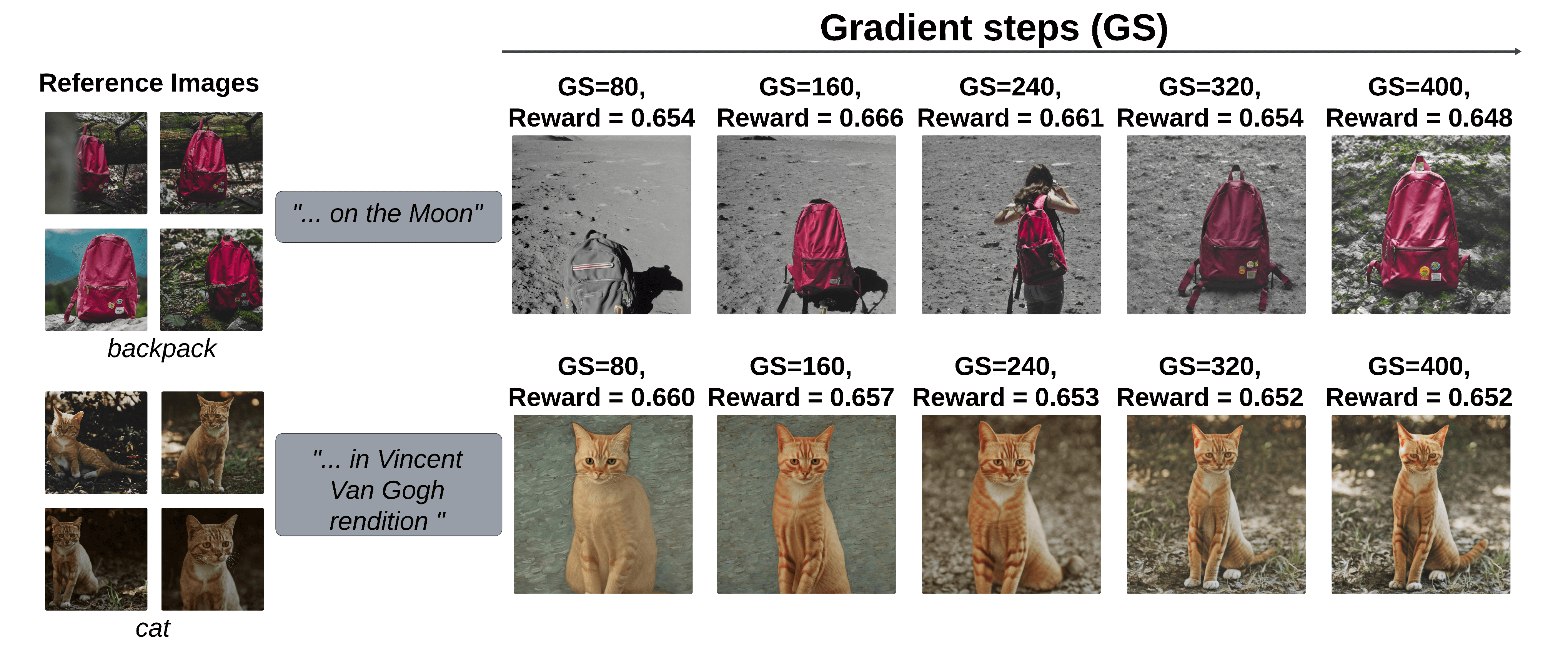}
    \caption{Changes in the $0.3$-Harmonic reward value during RPO training process.}
    \label{fig:reward_changing}
\end{figure}

\begin{figure}
    \centering
    \includegraphics[width=\textwidth]{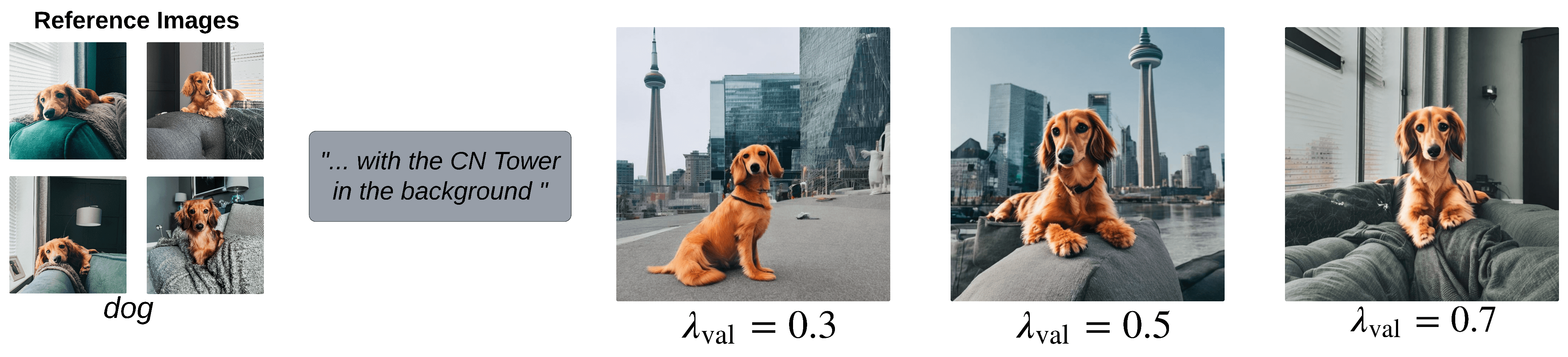}
    \caption{Different $\lambda_{\text{val}}$'s will lead to different results. A small $\lambda_{\text{val}}$ assigns a higher a weight for text-to-image alignment and leads to diverse generation. A large $\lambda_{\text{val}}$ may also cause overfitting.}
    \label{fig:lambda_comparison}
\end{figure}

\subsection{Ablation Study and Method Analysis}

\paragraph{Preference Loss and $\lambda$-Harmonic Ablation.} We investigate the second question through an ablation study. Two regularization components are introduced into RPO: reference loss as a regularizer and early stopping by $\lambda_{\text{val}}$-Harmonic reward function. Consequently, we compare four methods: (1) Pure $\Lc_{\text{sim}}$, which only minimizes the image-to-image similarity loss $\Lc_{\text{sim}}$; (2) $\Lc_{\text{pref}}$ w/o early stopping, which employs $\Lc_{\text{pref}}$ as a regularizer but omits early stopping by $\lambda_{\text{val}}$-Harmonic reward function; (3) Early stopping w/o $\Lc_{\text{pref}}$, which uses $\lambda_{\text{val}}$-Harmonic reward function as a regularization method but excludes $\Lc_{\text{pref}}$; (4) Full RPO, which utilizes both $\Lc_{\text{pref}}$ and early stopping by the $\lambda_{\text{val}}$-Harmonic reward function. We choose the default value $\lambda_{\text{val}} = 0.3$ in this ablation study.

Table~\ref{tab:ablation_study} lists the evaluation results of these four methods on DreamBench. We observe that without early stopping, $\Lc_{\text{pref}}$ can still prevent overfitting to the reference images and improve text-to-image alignment, though the regularization effect is weak. Specifically, the $0.3$-Harmonic only shows a marginal improvement of 0.003 over pure $\Lc_{\text{sim}}$ and 0.001 over early stopping without $\Lc_{\text{pref}}$. The early stopping facilitated by the $\lambda_{\text{val}}$-Harmonic reward function plays a crucial role in counteracting overfitting, helping the diffusion models retain the ability to generate high-quality images aligned with textual prompts. To provide a deeper understanding of the $\lambda$-Harmonic reward validation, we present two examples from during training in Figure \ref{fig:reward_changing}, covering both objects and live subjects. We found that the model tends to overfit at a very early stage, i.e., within 200 gradient steps, where $\lambda$-Harmonic can provide correct reward signals for the generated images. For the backpack subject, the generated image receives a low reward at gradient step 80 due to its lack of fidelity to the reference images. However, at gradient step 400, the image is overfitted to the reference images, and the model fails to align well with the input text, resulting in another low reward. $\lambda$-Harmonic assigns a high reward to images that are faithful to both the reference image and textual prompts.

\paragraph{Impact of $\lambda_{\text{val}}$.} We examine the third question by selecting different $\lambda_{\text{val}}$ values from the set $\{0.3, 0.5, 0.7\}$ as the validation parameters for the $\lambda$-Harmonic reward. According to Equation \ref{eq:method_reward_fn}, we believe that as $\lambda_{\text{val}}$ increases, the $\lambda$-Harmonic reward function will give higher weight to the image-to-image similarity score. This will make the generated images more closely resemble the reference images, however, there is also a risk of overfitting. Table~\ref{tab:lambda_comparison} shows the results of three different $\lambda_{\text{val}}$ values on DreamBench. As we expected, a larger $\lambda_{\text{val}}$ makes the images better preserve the characteristics of the reference images, but it also reduces the text-to-image alignment score. Figure~\ref{fig:lambda_comparison} shows us an example. In this example, different $\lambda_{\text{val}}$ values lead to different outcomes due to varying strengths of regularization. A smaller $\lambda_{\text{val}} = 0.3$ can generate more varied results, but seems somewhat off from the reference images. $\lambda_{\text{val}} = 0.5$ preserves some characteristics beyond the original subject, such as the sofa, but also maintains alignment between text and image. However, when $\lambda_{\text{val}} = 0.7$ is chosen as an excessively large value, the model actually overfits to the reference images, ignoring the prompts. We have additional comparisons, including different $\lambda_{\text{train}}$'s, different Pythagorean means, and aesthetic scores \cite{schuhmann2022laion} in Appendix~\ref{section:appendix_addition_comparison}.

\section{Conclusion}

We introduce the $\lambda$-Harmonic reward function to derive preference labels and employ RPO to finetune the diffusion model for subject-driven text-to-image generation tasks. Additionally, the $\lambda$-Harmonic reward function serves as a validation method, enabling early stopping to mitigate overfitting to reference images and speeding up the finetuning process. 

\begin{ack}
We thank Shixin Luo and Hongliang Fei for providing constructive feedback. This work was supported by a Google grant with Cloud TPUs from Google’s TPU Research Cloud (TRC).  We also thank the Vector Institute, the Canada CIFAR AI Chair program and the Natural
Sciences and Engineering Research Council of Canada for their support.
\end{ack}

\bibliographystyle{plainnat}
\bibliography{ref}

\begin{thebibliography}{33}
\providecommand{\natexlab}[1]{#1}
\providecommand{\url}[1]{\texttt{#1}}
\expandafter\ifx\csname urlstyle\endcsname\relax
  \providecommand{\doi}[1]{doi: #1}\else
  \providecommand{\doi}{doi: \begingroup \urlstyle{rm}\Url}\fi

\bibitem[Bai et~al.(2022)Bai, Jones, Ndousse, Askell, Chen, DasSarma, Drain,
  Fort, Ganguli, Henighan, et~al.]{bai2022training}
Yuntao Bai, Andy Jones, Kamal Ndousse, Amanda Askell, Anna Chen, Nova DasSarma,
  Dawn Drain, Stanislav Fort, Deep Ganguli, Tom Henighan, et~al.
\newblock Training a helpful and harmless assistant with reinforcement learning
  from human feedback.
\newblock \emph{arXiv preprint arXiv:2204.05862}, 2022.

\bibitem[Bird et~al.(2023)Bird, Ungless, and Kasirzadeh]{bird2023typology}
Charlotte Bird, Eddie Ungless, and Atoosa Kasirzadeh.
\newblock Typology of risks of generative text-to-image models.
\newblock In \emph{Proceedings of the 2023 AAAI/ACM Conference on AI, Ethics,
  and Society}, pages 396--410, 2023.

\bibitem[Black et~al.(2023)Black, Janner, Du, Kostrikov, and
  Levine]{black2023training}
Kevin Black, Michael Janner, Yilun Du, Ilya Kostrikov, and Sergey Levine.
\newblock Training diffusion models with reinforcement learning.
\newblock \emph{arXiv preprint arXiv:2305.13301}, 2023.

\bibitem[Bradley and Terry(1952)]{bradley1952rank}
Ralph~Allan Bradley and Milton~E Terry.
\newblock Rank analysis of incomplete block designs: I. the method of paired
  comparisons.
\newblock \emph{Biometrika}, 39\penalty0 (3/4):\penalty0 324--345, 1952.

\bibitem[Caron et~al.(2021)Caron, Touvron, Misra, J{\'e}gou, Mairal,
  Bojanowski, and Joulin]{caron2021emerging}
Mathilde Caron, Hugo Touvron, Ishan Misra, Herv{\'e} J{\'e}gou, Julien Mairal,
  Piotr Bojanowski, and Armand Joulin.
\newblock Emerging properties in self-supervised vision transformers.
\newblock In \emph{Proceedings of the IEEE/CVF international conference on
  computer vision}, pages 9650--9660, 2021.

\bibitem[Chen et~al.(2023)Chen, Zhang, Wu, Wang, Duan, Zhou, and
  Zhu]{chen2023disenbooth}
Hong Chen, Yipeng Zhang, Simin Wu, Xin Wang, Xuguang Duan, Yuwei Zhou, and
  Wenwu Zhu.
\newblock Disenbooth: Identity-preserving disentangled tuning for
  subject-driven text-to-image generation.
\newblock \emph{arXiv preprint arXiv:2305.03374}, 2023.

\bibitem[Chen et~al.(2022)Chen, Hu, Saharia, and Cohen]{chen2022re}
Wenhu Chen, Hexiang Hu, Chitwan Saharia, and William~W Cohen.
\newblock Re-imagen: Retrieval-augmented text-to-image generator.
\newblock \emph{arXiv preprint arXiv:2209.14491}, 2022.

\bibitem[Chen et~al.(2024)Chen, Hu, Li, Ruiz, Jia, Chang, and
  Cohen]{chen2024subject}
Wenhu Chen, Hexiang Hu, Yandong Li, Nataniel Ruiz, Xuhui Jia, Ming-Wei Chang,
  and William~W Cohen.
\newblock Subject-driven text-to-image generation via apprenticeship learning.
\newblock \emph{Advances in Neural Information Processing Systems}, 36, 2024.

\bibitem[Chowdhery et~al.(2023)Chowdhery, Narang, Devlin, Bosma, Mishra,
  Roberts, Barham, Chung, Sutton, Gehrmann, et~al.]{chowdhery2023palm}
Aakanksha Chowdhery, Sharan Narang, Jacob Devlin, Maarten Bosma, Gaurav Mishra,
  Adam Roberts, Paul Barham, Hyung~Won Chung, Charles Sutton, Sebastian
  Gehrmann, et~al.
\newblock Palm: Scaling language modeling with pathways.
\newblock \emph{Journal of Machine Learning Research}, 24\penalty0
  (240):\penalty0 1--113, 2023.

\bibitem[Djuki{\'c} et~al.(2006)Djuki{\'c}, Jankovi{\'c}, Mati{\'c}, and
  Petrovi{\'c}]{djukic2006imo}
Du{\v{s}}an Djuki{\'c}, Vladimir Jankovi{\'c}, Ivan Mati{\'c}, and Nikola
  Petrovi{\'c}.
\newblock \emph{The IMO compendium: a collection of problems suggested for the
  International Mathematical Olympiads: 1959-2004}, volume 119.
\newblock Springer, 2006.

\bibitem[Fan et~al.(2023)Fan, Watkins, Du, Liu, Ryu, Boutilier, Abbeel,
  Ghavamzadeh, Lee, and Lee]{fan2023reinforcement}
Ying Fan, Olivia Watkins, Yuqing Du, Hao Liu, Moonkyung Ryu, Craig Boutilier,
  Pieter Abbeel, Mohammad Ghavamzadeh, Kangwook Lee, and Kimin Lee.
\newblock Reinforcement learning for fine-tuning text-to-image diffusion
  models.
\newblock In \emph{Thirty-seventh Conference on Neural Information Processing
  Systems}, 2023.
\newblock URL \url{https://openreview.net/forum?id=8OTPepXzeh}.

\bibitem[Gal et~al.(2022)Gal, Alaluf, Atzmon, Patashnik, Bermano, Chechik, and
  Cohen-Or]{gal2022image}
Rinon Gal, Yuval Alaluf, Yuval Atzmon, Or~Patashnik, Amit~H Bermano, Gal
  Chechik, and Daniel Cohen-Or.
\newblock An image is worth one word: Personalizing text-to-image generation
  using textual inversion.
\newblock \emph{arXiv preprint arXiv:2208.01618}, 2022.

\bibitem[Ho et~al.(2020)Ho, Jain, and Abbeel]{ho2020denoising}
Jonathan Ho, Ajay Jain, and Pieter Abbeel.
\newblock Denoising diffusion probabilistic models.
\newblock \emph{Advances in neural information processing systems},
  33:\penalty0 6840--6851, 2020.

\bibitem[Jia et~al.(2021)Jia, Yang, Xia, Chen, Parekh, Pham, Le, Sung, Li, and
  Duerig]{jia2021scaling}
Chao Jia, Yinfei Yang, Ye~Xia, Yi-Ting Chen, Zarana Parekh, Hieu Pham, Quoc Le,
  Yun-Hsuan Sung, Zhen Li, and Tom Duerig.
\newblock Scaling up visual and vision-language representation learning with
  noisy text supervision.
\newblock In \emph{International conference on machine learning}, pages
  4904--4916. PMLR, 2021.

\bibitem[Kawar et~al.(2023)Kawar, Zada, Lang, Tov, Chang, Dekel, Mosseri, and
  Irani]{kawar2023imagic}
Bahjat Kawar, Shiran Zada, Oran Lang, Omer Tov, Huiwen Chang, Tali Dekel, Inbar
  Mosseri, and Michal Irani.
\newblock Imagic: Text-based real image editing with diffusion models.
\newblock In \emph{Proceedings of the IEEE/CVF Conference on Computer Vision
  and Pattern Recognition}, pages 6007--6017, 2023.

\bibitem[Kumari et~al.(2023)Kumari, Zhang, Zhang, Shechtman, and
  Zhu]{kumari2023multi}
Nupur Kumari, Bingliang Zhang, Richard Zhang, Eli Shechtman, and Jun-Yan Zhu.
\newblock Multi-concept customization of text-to-image diffusion.
\newblock In \emph{Proceedings of the IEEE/CVF Conference on Computer Vision
  and Pattern Recognition}, pages 1931--1941, 2023.

\bibitem[Loshchilov and Hutter(2017)]{loshchilov2017decoupled}
Ilya Loshchilov and Frank Hutter.
\newblock Decoupled weight decay regularization.
\newblock \emph{arXiv preprint arXiv:1711.05101}, 2017.

\bibitem[Ouyang et~al.(2022)Ouyang, Wu, Jiang, Almeida, Wainwright, Mishkin,
  Zhang, Agarwal, Slama, Ray, et~al.]{ouyang2022training}
Long Ouyang, Jeffrey Wu, Xu~Jiang, Diogo Almeida, Carroll Wainwright, Pamela
  Mishkin, Chong Zhang, Sandhini Agarwal, Katarina Slama, Alex Ray, et~al.
\newblock Training language models to follow instructions with human feedback.
\newblock \emph{Advances in neural information processing systems},
  35:\penalty0 27730--27744, 2022.

\bibitem[Radford et~al.(2021)Radford, Kim, Hallacy, Ramesh, Goh, Agarwal,
  Sastry, Askell, Mishkin, Clark, et~al.]{radford2021learning}
Alec Radford, Jong~Wook Kim, Chris Hallacy, Aditya Ramesh, Gabriel Goh,
  Sandhini Agarwal, Girish Sastry, Amanda Askell, Pamela Mishkin, Jack Clark,
  et~al.
\newblock Learning transferable visual models from natural language
  supervision.
\newblock In \emph{International conference on machine learning}, pages
  8748--8763. PMLR, 2021.

\bibitem[Rafailov et~al.(2024{\natexlab{a}})Rafailov, Hejna, Park, and
  Finn]{rafailov2024r}
Rafael Rafailov, Joey Hejna, Ryan Park, and Chelsea Finn.
\newblock From $ r $ to $ q^*$: Your language model is secretly a q-function.
\newblock \emph{arXiv preprint arXiv:2404.12358}, 2024{\natexlab{a}}.

\bibitem[Rafailov et~al.(2024{\natexlab{b}})Rafailov, Sharma, Mitchell,
  Manning, Ermon, and Finn]{rafailov2024direct}
Rafael Rafailov, Archit Sharma, Eric Mitchell, Christopher~D Manning, Stefano
  Ermon, and Chelsea Finn.
\newblock Direct preference optimization: Your language model is secretly a
  reward model.
\newblock \emph{Advances in Neural Information Processing Systems}, 36,
  2024{\natexlab{b}}.

\bibitem[Rombach et~al.(2022)Rombach, Blattmann, Lorenz, Esser, and
  Ommer]{rombach2022high}
Robin Rombach, Andreas Blattmann, Dominik Lorenz, Patrick Esser, and Bj{\"o}rn
  Ommer.
\newblock High-resolution image synthesis with latent diffusion models.
\newblock In \emph{Proceedings of the IEEE/CVF conference on computer vision
  and pattern recognition}, pages 10684--10695, 2022.

\bibitem[Ruiz et~al.(2023)Ruiz, Li, Jampani, Pritch, Rubinstein, and
  Aberman]{ruiz2023dreambooth}
Nataniel Ruiz, Yuanzhen Li, Varun Jampani, Yael Pritch, Michael Rubinstein, and
  Kfir Aberman.
\newblock Dreambooth: Fine tuning text-to-image diffusion models for
  subject-driven generation.
\newblock In \emph{Proceedings of the IEEE/CVF Conference on Computer Vision
  and Pattern Recognition}, pages 22500--22510, 2023.

\bibitem[Saharia et~al.(2022)Saharia, Chan, Saxena, Li, Whang, Denton,
  Ghasemipour, Gontijo~Lopes, Karagol~Ayan, Salimans,
  et~al.]{saharia2022photorealistic}
Chitwan Saharia, William Chan, Saurabh Saxena, Lala Li, Jay Whang, Emily~L
  Denton, Kamyar Ghasemipour, Raphael Gontijo~Lopes, Burcu Karagol~Ayan, Tim
  Salimans, et~al.
\newblock Photorealistic text-to-image diffusion models with deep language
  understanding.
\newblock \emph{Advances in neural information processing systems},
  35:\penalty0 36479--36494, 2022.

\bibitem[Schuhmann et~al.(2022)Schuhmann, Beaumont, Vencu, Gordon, Wightman,
  Cherti, Coombes, Katta, Mullis, Wortsman, et~al.]{schuhmann2022laion}
Christoph Schuhmann, Romain Beaumont, Richard Vencu, Cade Gordon, Ross
  Wightman, Mehdi Cherti, Theo Coombes, Aarush Katta, Clayton Mullis, Mitchell
  Wortsman, et~al.
\newblock Laion-5b: An open large-scale dataset for training next generation
  image-text models.
\newblock \emph{Advances in Neural Information Processing Systems},
  35:\penalty0 25278--25294, 2022.

\bibitem[Sohl-Dickstein et~al.(2015)Sohl-Dickstein, Weiss, Maheswaranathan, and
  Ganguli]{sohl2015deep}
Jascha Sohl-Dickstein, Eric Weiss, Niru Maheswaranathan, and Surya Ganguli.
\newblock Deep unsupervised learning using nonequilibrium thermodynamics.
\newblock In \emph{International conference on machine learning}, pages
  2256--2265. PMLR, 2015.

\bibitem[Song et~al.(2021)Song, Meng, and Ermon]{song2021denoising}
Jiaming Song, Chenlin Meng, and Stefano Ermon.
\newblock Denoising diffusion implicit models.
\newblock In \emph{International Conference on Learning Representations}, 2021.
\newblock URL \url{https://openreview.net/forum?id=St1giarCHLP}.

\bibitem[Song and Ermon(2019)]{song2019generative}
Yang Song and Stefano Ermon.
\newblock Generative modeling by estimating gradients of the data distribution.
\newblock \emph{Advances in neural information processing systems}, 32, 2019.

\bibitem[Song et~al.(2020)Song, Sohl-Dickstein, Kingma, Kumar, Ermon, and
  Poole]{song2020score}
Yang Song, Jascha Sohl-Dickstein, Diederik~P Kingma, Abhishek Kumar, Stefano
  Ermon, and Ben Poole.
\newblock Score-based generative modeling through stochastic differential
  equations.
\newblock \emph{arXiv preprint arXiv:2011.13456}, 2020.

\bibitem[Wallace et~al.(2023)Wallace, Dang, Rafailov, Zhou, Lou, Purushwalkam,
  Ermon, Xiong, Joty, and Naik]{wallace2023diffusion}
Bram Wallace, Meihua Dang, Rafael Rafailov, Linqi Zhou, Aaron Lou, Senthil
  Purushwalkam, Stefano Ermon, Caiming Xiong, Shafiq Joty, and Nikhil Naik.
\newblock Diffusion model alignment using direct preference optimization.
\newblock \emph{arXiv preprint arXiv:2311.12908}, 2023.

\bibitem[Wei et~al.(2023)Wei, Zhang, Ji, Bai, Zhang, and Zuo]{wei2023elite}
Yuxiang Wei, Yabo Zhang, Zhilong Ji, Jinfeng Bai, Lei Zhang, and Wangmeng Zuo.
\newblock Elite: Encoding visual concepts into textual embeddings for
  customized text-to-image generation.
\newblock In \emph{Proceedings of the IEEE/CVF International Conference on
  Computer Vision}, pages 15943--15953, 2023.

\bibitem[Ye et~al.(2023)Ye, Zhang, Liu, Han, and Yang]{ye2023ip}
Hu~Ye, Jun Zhang, Sibo Liu, Xiao Han, and Wei Yang.
\newblock Ip-adapter: Text compatible image prompt adapter for text-to-image
  diffusion models.
\newblock \emph{arXiv preprint arXiv:2308.06721}, 2023.

\bibitem[Zhang et~al.(2024)Zhang, Song, Liu, Wang, Yu, Tang, Li, Tang, Hu, Pan,
  et~al.]{zhang2024ssr}
Yuxuan Zhang, Yiren Song, Jiaming Liu, Rui Wang, Jinpeng Yu, Hao Tang, Huaxia
  Li, Xu~Tang, Yao Hu, Han Pan, et~al.
\newblock Ssr-encoder: Encoding selective subject representation for
  subject-driven generation.
\newblock In \emph{Proceedings of the IEEE/CVF Conference on Computer Vision
  and Pattern Recognition}, pages 8069--8078, 2024.

\end{thebibliography}

\clearpage

\appendix

\section{Appendix}

\subsection{Background}\label{section:appendix_background}

\paragraph{Reinforcement Learning.} In Reinforcement Learning (RL), the environment can be formalized as a Markov Decision Process (MDP). An MDP is defined by a tuple $(\Sc, \Ac, P, R, \rho_0, T)$, where $\Sc$ is the state space, $\Ac$ is the action space, $P$ is the transition function, $R$ is the reward function, $\rho_0$ is the distribution over initial states, and $T$ is the time horizon. At each timestep $t$, the agent observes a state $\sv_t$ and selects an action $\av_t$ according to a policy $\pi(\av_t | \sv_t)$, and  obtains a reward $R(\sv_t, \av_t)$, and then transit to a next state $\sv_{t + 1} \sim P(\sv_{t + 1} | \sv_t, \av_t).$ As the agent interacts with the MDP, it produces a sequence of states and actions, which is denoted as a trajectory $\tauv = (\sv_0, \av_0, \sv_1, \av_1, \dots, \sv_{T - 1}, \av_{T - 1})$. The RL objective is to maximize the expected value of cumulative reward over the trajectories sample from its policy:
\begin{align*}
    \Eb_{\tauv \sim p^\pi(\tauv)}\left[\sum_{t=0}^{T - 1} R(\sv_t, \av_t)\right]
\end{align*}

\paragraph{Diffusion models.} \cite{sohl2015deep, ho2020denoising, song2020score} consider a variance schedule $\{\beta_t\}_{t = 1}^T$. For each data point $\xv_0 \sim p_{\text{data}}(\xv_0)$, a Markov chain is constructed by transition probability $p(\xv_t \mid \xv_{t - 1}) = \Nc(\xv_t; \sqrt{1 - \beta_t} \xv_{t - 1}, \beta_t \Iv)$. Consequently, $p_{\alpha_t}(\xv_t \mid \xv_0) = \Nc(\xv_t; \sqrt{\alpha_t} \xv_0, (1 - \alpha_t) \Iv), $ where $\alpha_t = \prod_{t^\prime = 1}^t(1 - \beta_t)$. \cite{song2019generative, song2020score} show that training diffusion models, $\epsilonv_{\thetav}(\xv_t, t)$, is equivalent to score matching, i.e., training a score function $\sv_{\thetav}(\xv_t, t)$:
\begin{align}
    & \Eb_{t \sim \Uc\{1, \dots, T\}, \xv_0 \sim p_{\text{data}}, \xv_t \sim p_{\alpha_t}(\xv_t \mid \xv_0)}\left[\lVert \nabla_{\xv_t} \log p_{\alpha_t}(\xv_t \mid \xv_0) - \sv_{\thetav}(\xv_t, t)\rVert_2^2\right] \nonumber \\
    = & \Eb_{t \sim \Uc\{1, \dots, T\}, \xv_0 \sim p_{\text{data}}, \xv_t \sim p_{\alpha_t}(\xv_t \mid \xv_0)}\left[\left\lVert -\frac{1}{1 - \alpha_t} (\xv_t - \sqrt{\alpha_t} \xv_0) - \sv_{\thetav}(\xv_t, t)\right\rVert_2^2 \right] \nonumber \\
    = & \Eb_{t \sim \Uc\{1, \dots, T\}, \xv_0 \sim p_{\text{data}}, \epsilonv \sim \Nc(\mathbf{0}, \Iv)} \left[\left\lVert -\sqrt{1 - \alpha_t} \epsilonv - \sv_{\thetav}(\xv_t, t)\right\rVert_2^2\right] \because \xv_t = \sqrt{\alpha_t} \xv_0 + \sqrt{1 - \alpha_t} \epsilonv \nonumber \\
    = & \Eb_{t \sim \Uc\{1, \dots, T\}, \xv_0 \sim p_{\text{data}}, \epsilonv \sim \Nc(\mathbf{0}, \Iv)} \left[(1 - \alpha_t)\lVert \epsilonv_{\thetav}(\xv_t, t) - \epsilonv\rVert_2^2\right] \quad \text{let } \epsilonv_{\thetav}(\xv_t, t) \coloneqq -\frac{\sv_{\thetav}(\xv_t, t)}{\sqrt{1 - \alpha_t}} \nonumber \\
    = & \Eb_{t \sim \Uc\{1, \dots, T\}, \xv_0 \sim p_{\text{data}}, \epsilonv \sim \Nc(\mathbf{0}, \Iv)} \left[\omega(t)\lVert \epsilonv_{\thetav}(\xv_t, t) - \epsilonv\rVert_2^2\right], \quad \text{where } \omega(t) \coloneqq 1 - \alpha_t. \nonumber
\end{align}

\paragraph{Diffusion MDP} We formalize the denoising process as the following \textit{Diffusion MDP}:
\begin{align*}
    &\sv_{T - t} = (\xv_t, \cv), \quad \av_{T - t} = \xv_{t - 1}, \quad \pi_{\thetav}(\av_{T - t} \mid \sv_{T - t}) = \ptheta(\xv_{t - 1} \mid \xv_t, \cv), \\
    & \rho_0 = p(\cv) \times \Nc(\mathbf{0}, \Iv), \quad R(\sv_{T - t}, \av_{T - t}) = R(\xv_t, \xv_{t-1}, \cv)= \begin{cases}
    r(\xv_0, \cv) &\text{ if } t = 1, \\
    0 & \text{otherwise}
    \end{cases}
\end{align*}
where $r(\xv_0, \cv)$ can be a reward signal for the denoised image. The transition kernel is deterministic, i.e.,  $P(\sv_{T - t + 1} \mid \sv_{T - t}, \av_{T - t}) = (\av_{T - t}, \cv) = (\xv_{t - 1}, \cv).$ For brevity, the trajectory $\tauv$ is defined by $(\xv_0, \xv_1, \dots, \xv_T).$ Hence, the trajectory distribution for given diffusion models can be denoted as the joint distribution $\ptheta(\xv_{0:T} \mid \cv)$. In particular, the RL objective function for finetuning diffusion models can be re-written as the following optimization problem:
\begin{align*}
    \Eb_{\xv_{0:T} \sim \ptheta(\xv_{0:T} \mid \cv)}\left[\sum_{t=1}^{T} R(\xv_t, \xv_{t-1}, \cv) - \beta \Db_{\text{KL}}(\ptheta(\xv_{t - 1} \mid \xv_t, \cv) \| \pbase(\xv_{t - 1}\mid \xv_t, \cv))\right],
\end{align*}
where $\beta$ is a hyperparameter controlling the KL-divergence between the finetune model $\ptheta$ and the base model $\pbase$. This constraint prevents the learned model from losing generation diversity and falling into 'mode collapse' due to a single high cumulative reward result. In practice, this KL-divergence has become a standard constraint in large language model finetuning \citep{bai2022training, ouyang2022training, rafailov2024direct}. 

Given preference labels and following the Direct Preference Optimization (DPO) framework \cite{wallace2023diffusion, rafailov2024direct, rafailov2024r}, we can approximate the optimal policy $p^*$ by minimizing the upper bound:
\begin{align*}
    \Eb_{\xv^+_0, \xv^-_0, t, \xv^+_t, \xv^-_t} &\Bigg[-\log \sigma \bigg(\beta \big(\lVert \epsilonv_{\text{base}}(\xv^+_t, \cv, t) - \epsilonv^+_t \rVert^2_2 - \lVert \epsilonv_{\thetav}(\xv^+_t, \cv , t) - \epsilonv^+ \rVert^2_2  \nonumber \\ 
    & - (\lVert \epsilonv_{\text{base}}(\xv^-_t, \cv, t) - \epsilonv^-_t \rVert^2_2 - \lVert \epsilonv_{\thetav}(\xv^-_t, \cv , t) - \epsilonv^- \rVert^2_2) \big)\bigg)\Bigg],
\end{align*}
where $\epsilonv^+$ and $\epsilonv^-$ are independent samples from a Gaussian distribution, $\xv^+_t$ and $\xv^-_t$ are perturbed versions of $\xv^+_0$ and $\xv^-_t$ that depend on $\epsilonv^+$ and $\epsilonv^-$, and $\xv^+_0$ is preferred to $\xv^-_0$. A detailed derivation can be found in \cite{wallace2023diffusion}.

\subsection{Experimental Details} \label{section:appendix_experiments_details}

\begin{figure}
    \centering
    \includegraphics[width=\textwidth]{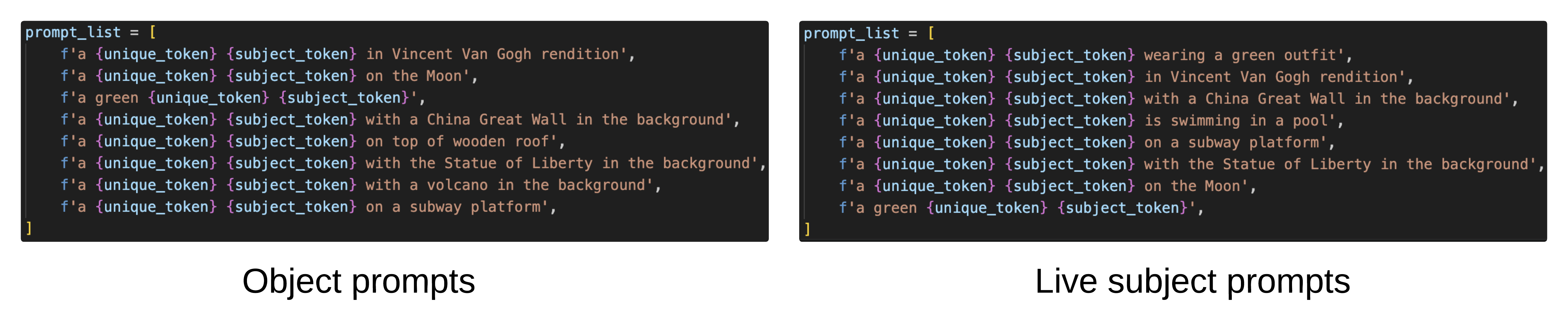}
    \caption{Training prompts for objects and live subjects.}
    \label{fig:appendix_training_prompts}
\end{figure}

\paragraph{Training Prompts.} We collect 8 training prompts: 6 re-contextualization, 1 property modification and 1 artistic style transfer for objects. 5 re-contextualization, 1 attribute editing, 1 artistic style transfer and 1 accessorization for live subjects. The trainig prompts are shown in Figure \ref{fig:appendix_training_prompts}.

\paragraph{Experimental Setup} During training we use $\lambda_{\text{train}} = 0$ for the $\lambda$-Harmonic reward function to generate the preference labels. We evaluate the model performance by $\lambda_{\text{val}}$-Harmonic per 40 gradient steps during training time and save the checkpoint that achieve the highest validation reward. Table \ref{tab:hyperparameters} lists the common hyperparameters used in the generating skill set and the $\lambda_{\text{val}}$ used in the default setting.

\begin{table}
  \caption{RPO Hyperparameters}
  \label{tab:hyperparameters}
  \centering
  \begin{tabular}{l|l}
    \toprule
    Parameter & Value \\
    \midrule
    \textit{Optimization} \\
    \quad optimizer & AdamW \cite{loshchilov2017decoupled} \\
    \quad learning rate & $5 \times 10^{-6}$ \\
    \quad weight decay & $0.01$ \\
    \quad gradient clip norm & $1.0$ \\
    \midrule
    \textit{RPO} \\
    \quad regularizer weight, $\beta$ & $1.0$ \\
    \quad gradient steps & $400$ \\
    \quad training preference weight, $\lambda_{\text{train}}$ & $0.0$ \\
    \quad validation preference weight (default), $\lambda_{\text{val}}$ & $0.3$ \\
    \bottomrule
  \end{tabular}
\end{table}

\subsection{Additional Comparisons}\label{section:appendix_addition_comparison}

\begin{figure}
    \centering
    \includegraphics[width=\textwidth]{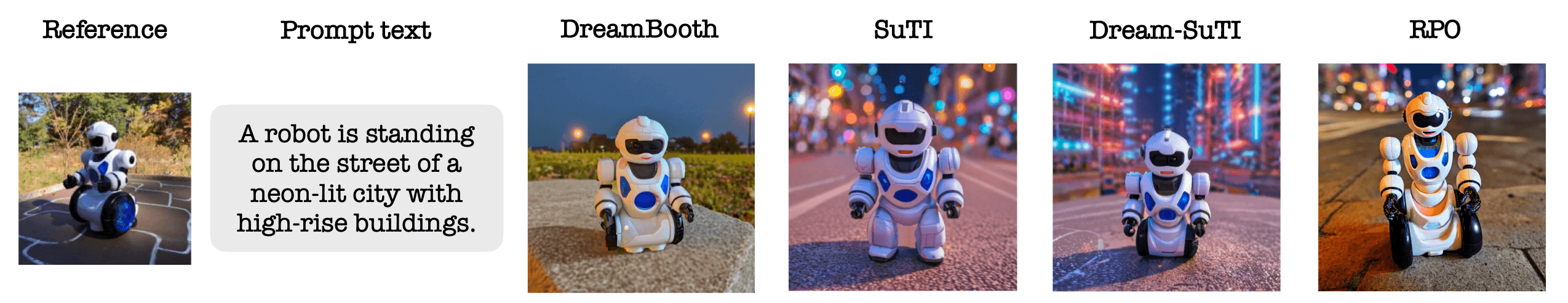}
    \caption{Comparison between DreamBooth, SuTI, Dream-SuTI, and RPO, adapted from \cite{chen2024subject}}
    \label{fig:appendix_suti_fail}
\end{figure}

\paragraph{Comparison to DreamBooth and SuTI.} We observe RPO that can be faithful to both reference images and the input prompt. To investigate whether RPO can provide better quality than DreamBooth and SuTI, we follow SuTI paper and pick the robot toy as an example to demonstrate the effectiveness of RPO (Figure \ref{fig:appendix_suti_fail}). In this example, DreamBooth is faithful to the reference image but it does not provide a good text-to-image alignment. SuTI provides an result that is fidelty to textual prompt. However, SuTI lacks fidelity to the reference image, i.e., the robot should stand with its wheels instead of legs. \cite{chen2024subject} use DreamBooth to finetune SuTI (Dream-SuTI) further to solve this failure case. Instead, RPO can generate an image not only faithful to the reference images but also align well with the input prompts.

\begin{figure}
    \centering
    \includegraphics[width=0.9\textwidth]{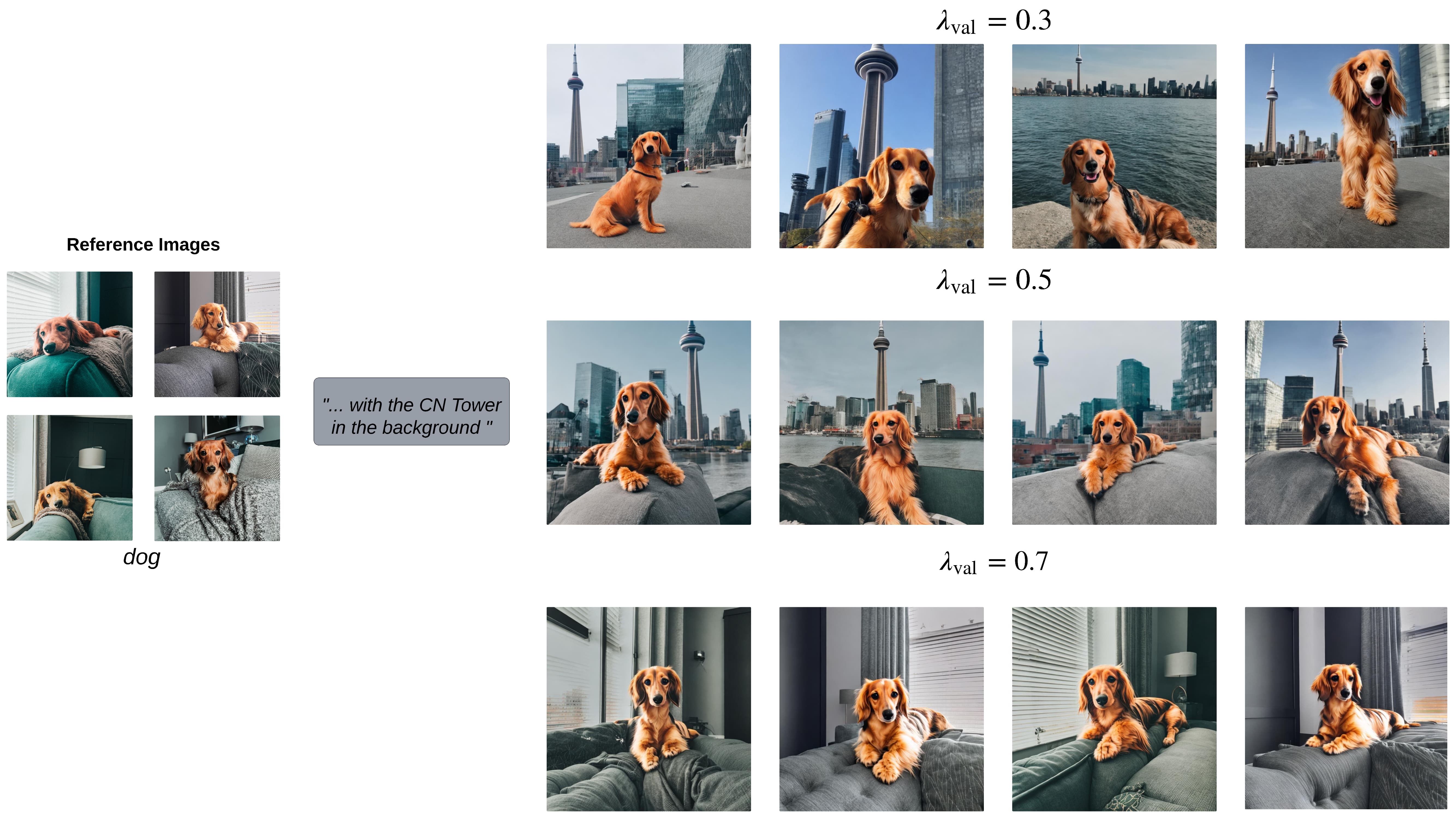}
    \caption{Additional samples of different $\lambda_{\text{val}}$. A small $\lambda_{\text{val}}$ leads to more diverse generated images.}
    \label{fig:appendix_additional_comparison}
\end{figure}

\paragraph{Comparison to different $\lambda_{\text{val}}$.} We have also added more samples for comparison of different $\lambda_{\text{val}}$ values (see Figure \ref{fig:appendix_additional_comparison}). We find that $\lambda_{\text{val}} = 0.5$ encourages the learned model to retain output diversity while still aligning with textual prompts. However, the generated images invariably contain a sofa, which is unrelated to the subject images. This occurs because every training image includes a sofa. A large $\lambda_{\text{val}}$ weakens the regularization strength and leads to overfitting. Nevertheless, a small value of $\lambda_{\text{val}}$ can potentially eliminate background bias. We highlight that this small $\lambda_{\text{val}}$ not only encourages diversity but also mitigates background bias in identity preservation and enables the model to focus on the subject.

\paragraph{Comparison to different $\lambda_{\text{train}}$.} During our experiments, we use the default value for $\lambda_{\text{train}} = 0$. Thus, we also investigate how different $\lambda_{\text{train}}$ will effect the performance of RPO. We test three different values of $\lambda_{\text{train}} = \{0.3, 0.5, 0.7\}$ and report their results in DreamBench (Table \ref{tab:lambda_train_comparison}). We observe that the image-to-image alignment increases with larger , but the text-to-image alignment decreases because the preference model tends to favor alignment with reference images and ignores the prompt alignment.

\paragraph{Comparison to different Pythagorean means.} Further, we examine different Pythagorean means of ALIGN-I and ALIGN-T scores. We replace the harmonic mean as the arithmetic mean and test three different $\lambda_{\text{val}} = \{0.3, 0.5, 0.7\}$. We report the results of the arithmetic mean reward in Table \ref{tab:arithmetic_mean_comparison}. The arithmetic mean is not very sensitive to smaller values; it tends to maximize higher values to achieve a better final score. In practice, ALIGN-I will receive a higher value (this effect can be seen from CLIP-I and CLIP-T in Tables 1 to 3). Thus, the model will tend to optimize image-to-image alignment and achieve good results on DINO and CLIP-I but have a lower score for text-to-image alignment.

\paragraph{Comparison to DPO} The original DPO is not suitable for subject-driven tasks because the datasets do not contain preference labels. We introduce the $\lambda$-harmonic function and design a variant of DPO for this task. We implement a DPO diffusion \cite{wallace2023diffusion} (without similarity loss) using preference labels for image-to-image similarity and text-to-image alignment. We choose $\lambda_{\text{train}} = 0.5$ since this value assigns equal weights to the image-to-image similarity and text-to-image alignment. For a fair comparison, we also repost the results from RPO with the same $\lambda_{\text{train}}$ value. The results on DreamBench for these two methods are shown in Table \ref{tab:dpo_comparison}. These results show that DPO can capture the text-to-image alignment from the preference labels. However, without $\Lc_{\text{sim}}$, DPO faces a significant overfitting problem; i.e., it achieves high text-to-image alignment but cannot preserve the subject's unique features.

\paragraph{Aesthetic comparison.} Limited evaluation metrics are a common issue in subject-driven tasks. In Table \ref{tab:aethestic_score}, we report the average aesthetic scores \cite{schuhmann2022laion} of the real reference images in DreamBench and the average aesthetic scores obtained with the best CLIP I/T $\lambda$ configuration ($\lambda_{\text{val}} = 0.5$) in DreamBench. We observe that RPO does not decrease the quality of images. Instead, the generated images achieve slightly better quality than the reference images.

\begin{table}
  \caption{Different $\lambda_{\text{train}}$-Harmonic reward comparison for evaluating fidelity over multiple subjects and prompts.}
  \label{tab:lambda_train_comparison}
  \centering
  \begin{tabular}{lllll}
    \toprule
    Configuration & DINO $\uparrow$ & CLIP-I $\uparrow$ & CLIP-T $\uparrow$ \\
    \midrule
    $\lambda_{\text{train}} = 0.0$ (default) & $0.581 \pm 0.113$ & $0.798 \pm 0.039$ & $\bm{0.329 \pm 0.021}$ \\
    $\lambda_{\text{train}} = 0.3$ & $0.646 \pm 0.083$  & $0.815 \pm 0.037$ & $0.315 \pm 0.026$ \\
    $\lambda_{\text{train}} = 0.5$ & $0.649 \pm 0.080$ & $0.829 \pm 0.039$ & $0.314 \pm 0.026$ \\
    $\lambda_{\text{train}} = 0.7$ & $\bm{0.651 \pm 0.088}$ & $\bm{0.831 \pm 0.033}$ & $0.314 \pm 0.026$ \\
    \bottomrule
  \end{tabular}
\end{table}

\begin{table}
  \caption{Comparison for harmonic mean and arithmetic mean reward function.}
  \label{tab:arithmetic_mean_comparison}
  \centering
  \begin{tabular}{lllll}
    \toprule
    Configuration & DINO $\uparrow$ & CLIP-I $\uparrow$ & CLIP-T $\uparrow$ \\
    \midrule
    $\lambda_{\text{val}} = 0.3$ (arithmetic mean) & $0.638 \pm 0.083$ & $0.823 \pm 0.037$ & $0.318 \pm 0.027$ \\
    $\lambda_{\text{val}} = 0.5$ (arithmetic mean) & $\bm{0.702 \pm 0.078}$  & $\bm{0.857 \pm 0.047}$ & $0.295 \pm 0.017$ \\
    $\lambda_{\text{val}} = 0.7$ (arithmetic mean) & $0.678 \pm 0.085$ & $0.851 \pm 0.041$ & $0.299 \pm 0.026$ \\
    \midrule
    $\lambda_{\text{val}} = 0.3$ (harmonic mean) & $0.581 \pm 0.113$ & $0.798 \pm 0.039$ & $\bm{0.329 \pm 0.021}$ \\
    $\lambda_{\text{val}} = 0.5$ (harmonic mean) & $0.652 \pm 0.082$ & $0.833 \pm 0.041$ & $0.314 \pm 0.022$ \\
    $\lambda_{\text{val}} = 0.7$ (harmonic mean) & $0.679 \pm 0.085$ & $0.850 \pm 0.045$ & $0.304 \pm 0.023$ \\
    \bottomrule
  \end{tabular}
\end{table}

\begin{table}
    \caption{DPO comparison for evaluating fidelity over multiple subjects and prompts on DreamBench}
    \label{tab:dpo_comparison}
    \centering
    \begin{tabular}{llll}
          \toprule
          Method & DINO $\uparrow$ & CLIP-I $\uparrow$ & CLIP-T $\uparrow$ \\
          \midrule
          DPO & $0.338$ & $0.702$ & $\bm{0.334}$ \\
          \midrule
          Ours: RPO & $\bm{0.649}$ & $0.819$ & $0.314$ \\
          \bottomrule
    \end{tabular}
\end{table}

\begin{table}
    \caption{Aesthetic score comparison for the real reference images and RPO generated images}
    \label{tab:aethestic_score}
    \centering
    \begin{tabular}{ll}
          \toprule
          Method & Aesthetic Score $\uparrow$ \\
          \midrule
          Real images & $5.145 \pm 0.312$ \\
          \midrule
          Ours: RPO ($\lambda_{\text{val}} = 0.5$)& $\bm{5.208 \pm 0.327}$ \\
          \bottomrule
    \end{tabular}
\end{table}

\subsection{Limitations}\label{section:appendix_limitations}

\begin{figure}[h]
    \centering
    \includegraphics[width=0.8\textwidth]{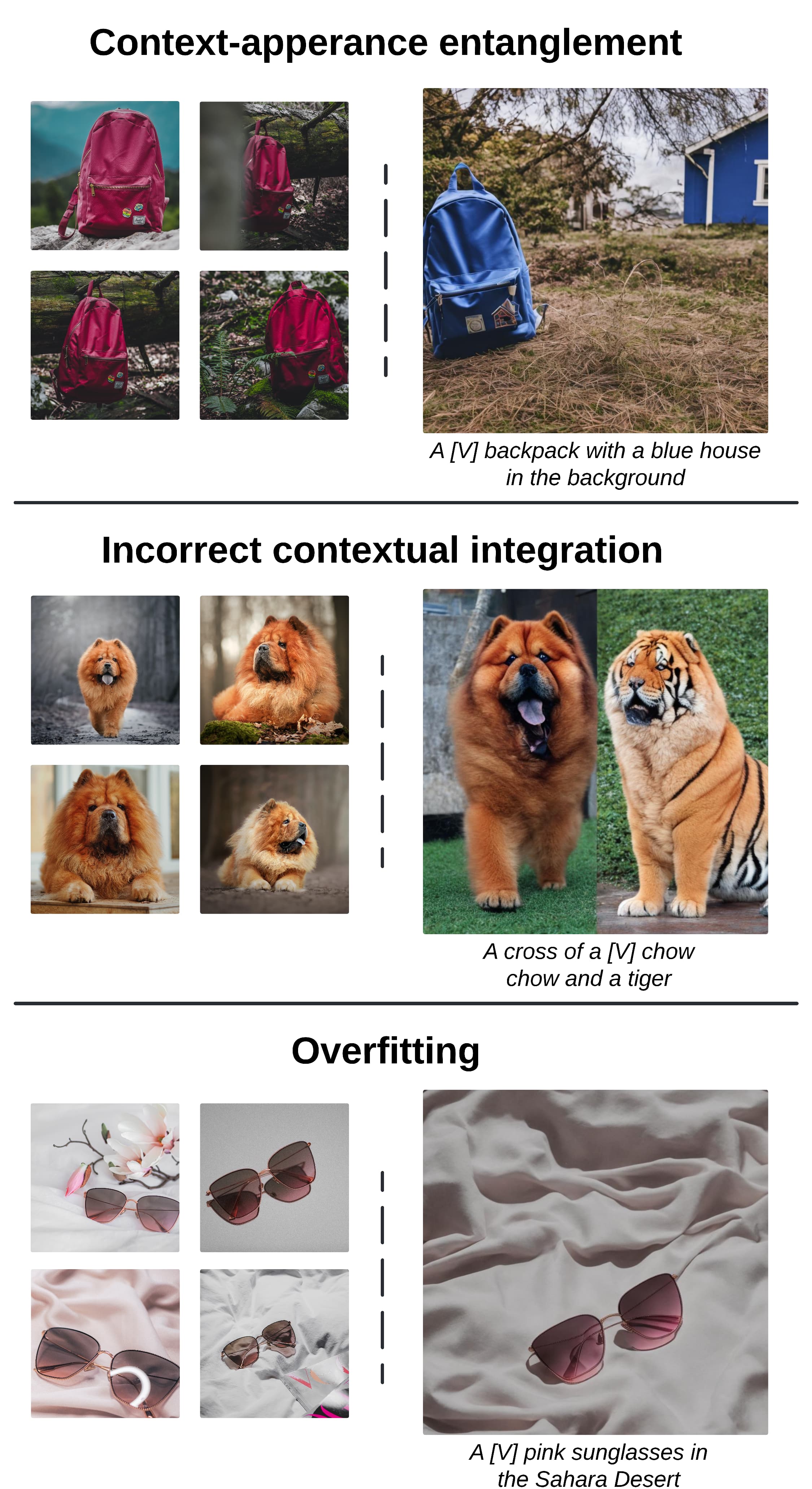}
    \caption{RPO's failure modes include: (1) The context and the appearance of the subject becoming entangled. (2) The trained model failing to generate an image with respect to the input prompt. (3) The trained model still overfitting to the training set.}
    \label{fig:failure_case}
\end{figure}

Figure~\ref{fig:failure_case} illustrates some failure examples of RPO. The first issue is context-appearance entanglement. In Figure \ref{fig:failure_case}, the learned model correctly understands the keyword \textit{blue house}; however, the appearance of the subject is also altered by this context, e.g. the color of the backpack has changed, and there is a house pattern on the backpack. The second issue is incorrect contextual integration. We conjecture that this failure is due to the rarity of the textual prompt. For instance, imagining a cross between a chow chow and a tiger is challenging, even for humans. Third, although RPO provides regularization, it still cannot guarantee the avoidance of overfitting. As shown in Figure~\ref{fig:failure_case}, this may be because, to some extent, the visual appearance of sand and bed sheets is similar, which has led to overfitting issues in the model. 

\subsection{Future Work}\label{section:appendix_future_work}
The overfitting failure case leads to a future work direction: \textit{can online RL improve regularization and avoid overfitting?} The second direction for future work involves implementing the LoRA version for RPO and comparing it to LoRA DreamBooth. Last but not least, we aim to identify or construct open-source, subject-driven datasets for comparison. Currently, DreamBench is the only open-source dataset we can access and evaluate for model performance. Nevertheless, we should create a larger dataset that includes more diverse subjects to verify the effectiveness of different algorithms.

\subsection{Broader Impacts}\label{appendix:broaderimpacts}

The nature of generative image models is inherently subjected to criticism on the issue of privacy, security and ethics in the presence of nefarious actors. However, the core of this paper remains purely on an academic mission to extend the boundaries of generative models. The societal consequences of democratizing such powerful generative models is more thoroughly discussed in other papers. For example, Bird et al. outlines the classes of risks in text-to-image models \cite{bird2023typology} in greater detail and should be directed to such papers. Nevertheless, we play our part in the management of such risks by avoiding the use of identifiable parts of humans in the reference sets.

\subsection{Skill Set}\label{section:appendix_skillset}

\begin{figure}
    \centering
    \includegraphics[width=\textwidth]{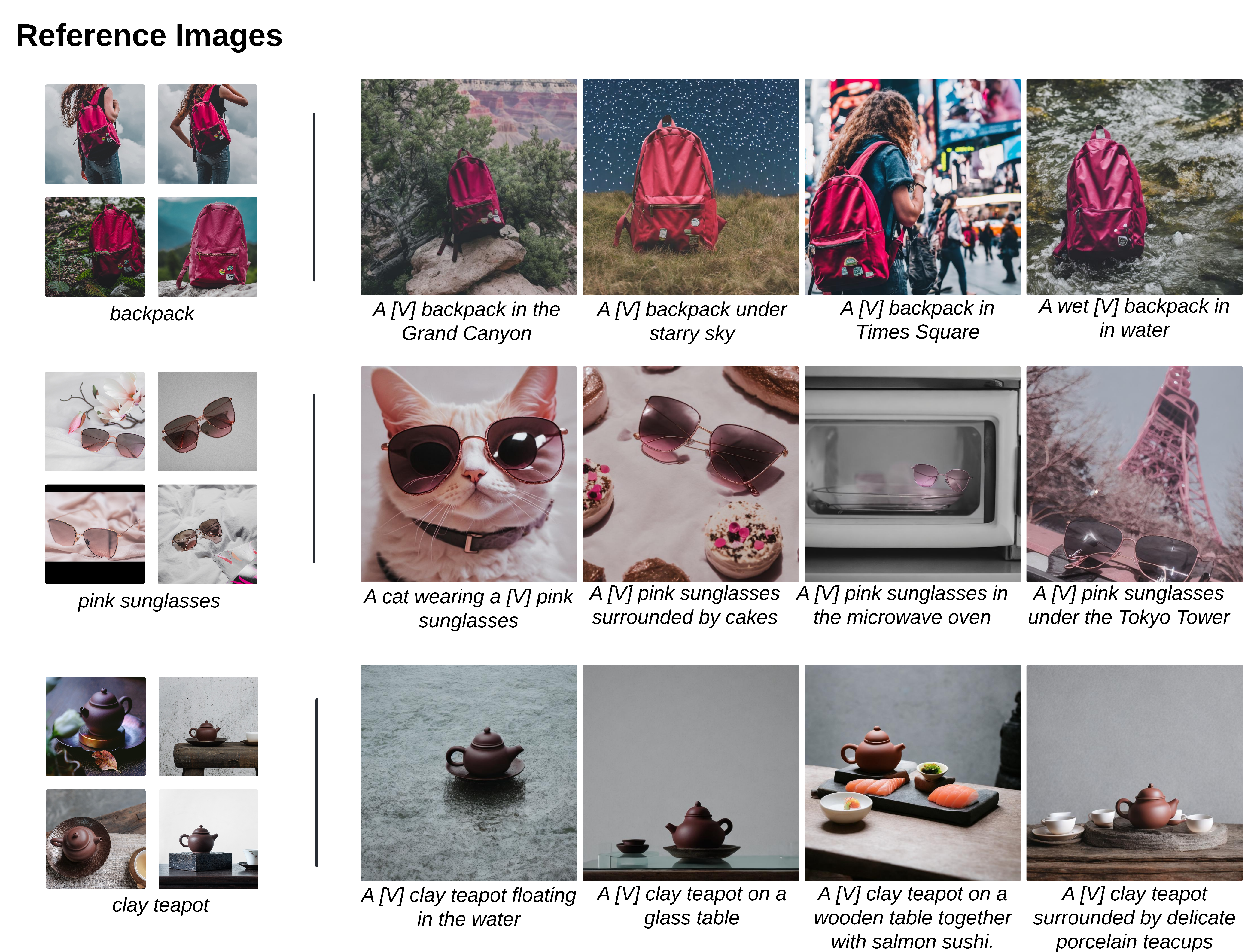}
    \caption{Re-contextualization samples from the RPO algorithm. RPO is able to generate images of specific subjects in unseen environments while preserving the identity and details of the subjects and be faithful to the input prompts. Reference images are shown on the left. We display the generated images along with their textual prompts on the right.}
    \label{fig:recontext}
\end{figure}

\begin{figure}
    \centering
    \includegraphics[width=\textwidth]{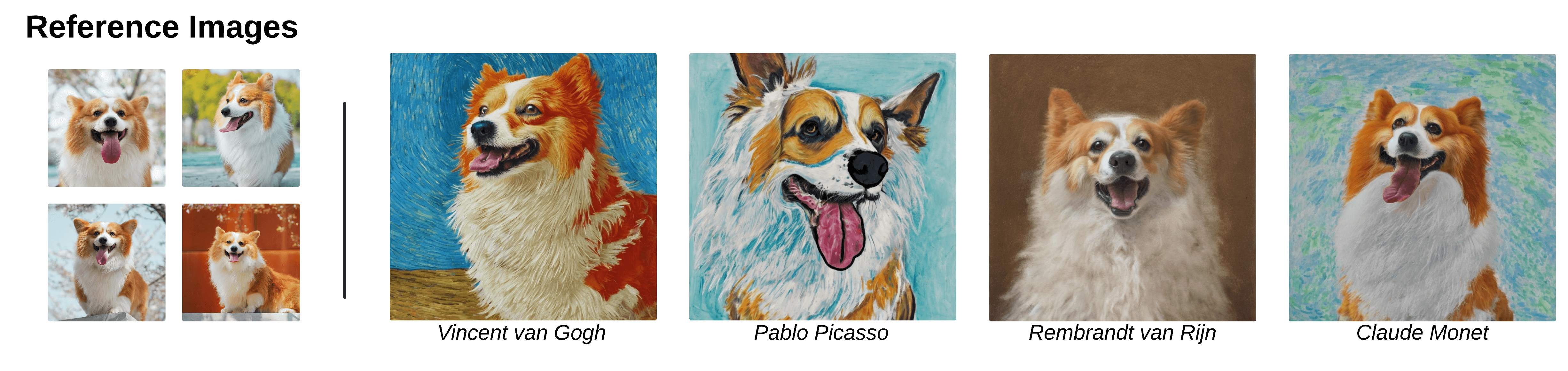}
    \caption{Artistic style rendition samples from the RPO algorithm. These art renditions can be applied to a dog given a prompt of \texttt{``a [painter] styled painting of a [V] dog''}. The identity of the subject is preserved and faithfully imitates the style of famous painters.}
    \label{fig:art_renditions}
\end{figure}

\begin{figure}
    \centering
    \includegraphics[width=\textwidth]{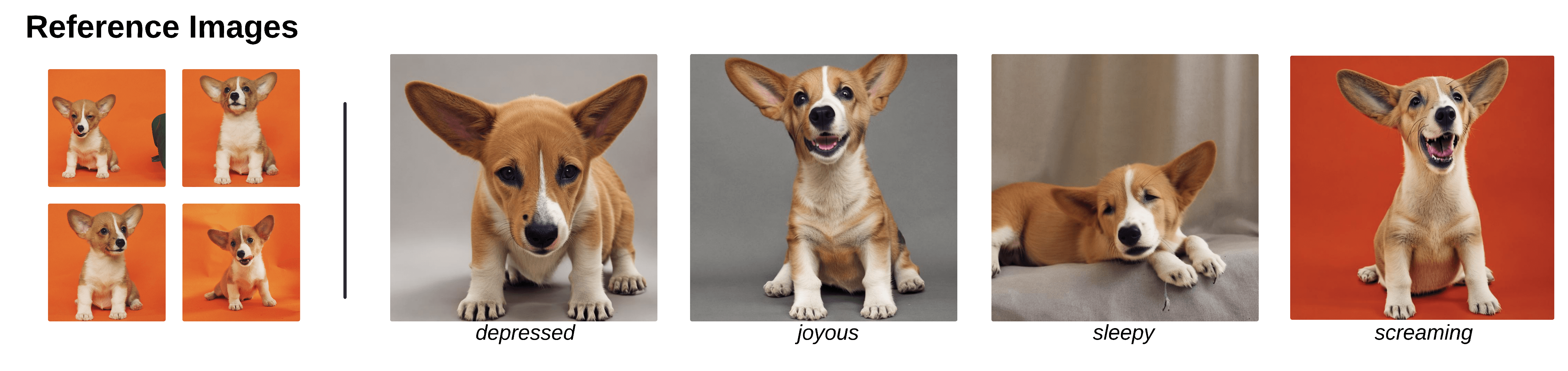}
    \caption{Expression modification samples from the RPO algorithm. RPO can integrate subject with various unseen expressions in the reference images. We also notice the pose of generated images, e.g., \texttt{a [V] sleepy dog}, were not displayed in the training set.}
    \label{fig:expression_modification}
\end{figure}

\begin{figure}
    \centering
    \includegraphics[width=\textwidth]{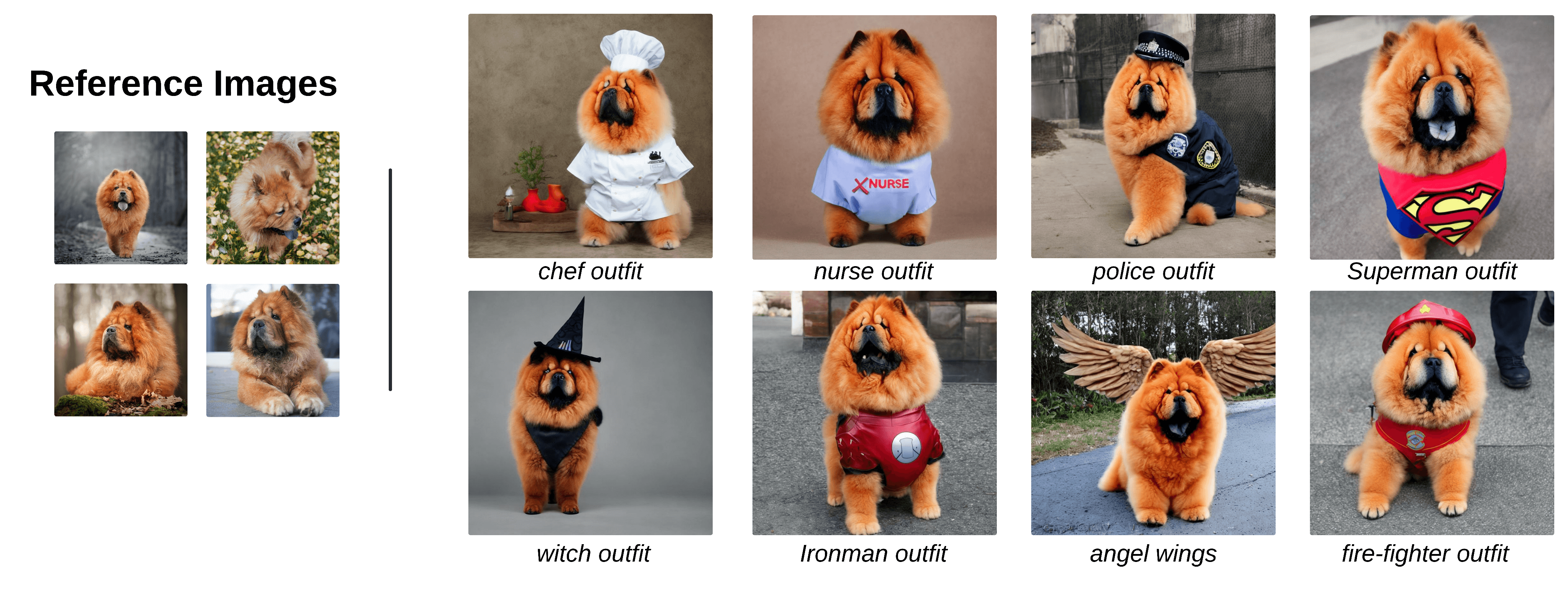}
    \caption{Accessories samples from the RPO algorithm. Conditioned the prompts, \texttt{``a [V] chow chow wearing a [target outfit]''}, the generated images retains the unique features of the reference images, e.g., the hair color and breed of the subject dog. The interaction between subject dog and the outfits is realistic.}
    \label{fig:outfig_accessories}
\end{figure}

\begin{figure}
    \centering
    \includegraphics[width=\textwidth]{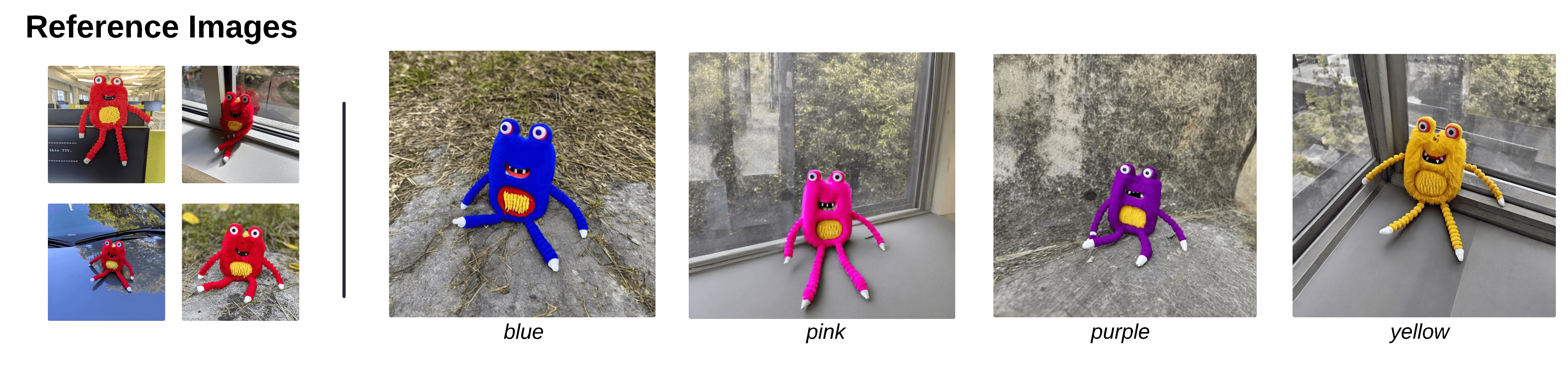}
    \caption{Color editing samples from the RPO algorithm. We display color modifications using prompts such as \texttt{``a [target color] [V] monster toy''}. The identity of the subject is preserved.}
    \label{fig:color_editing}
\end{figure}

\begin{figure}
    \centering
    \includegraphics[width=\textwidth]{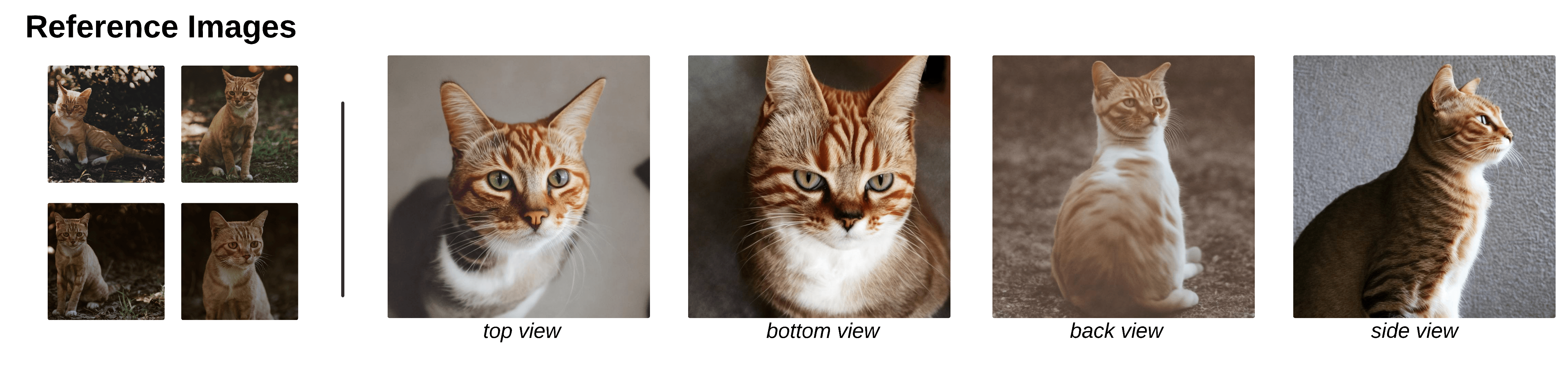}
    \caption{Multi-view samples from the RPO algorithm. We generate images from specified viewpoints of the subject. For the top and bottom views, we use the prompts \texttt{``a [V] cat looking up/down at the camera''}. For the back and side views, we apply the prompts \texttt{``a back/side view of a [V] cat''}.}
    \label{fig:multiple_view}
\end{figure}

\begin{figure}
    \centering
    \includegraphics[width=\textwidth]{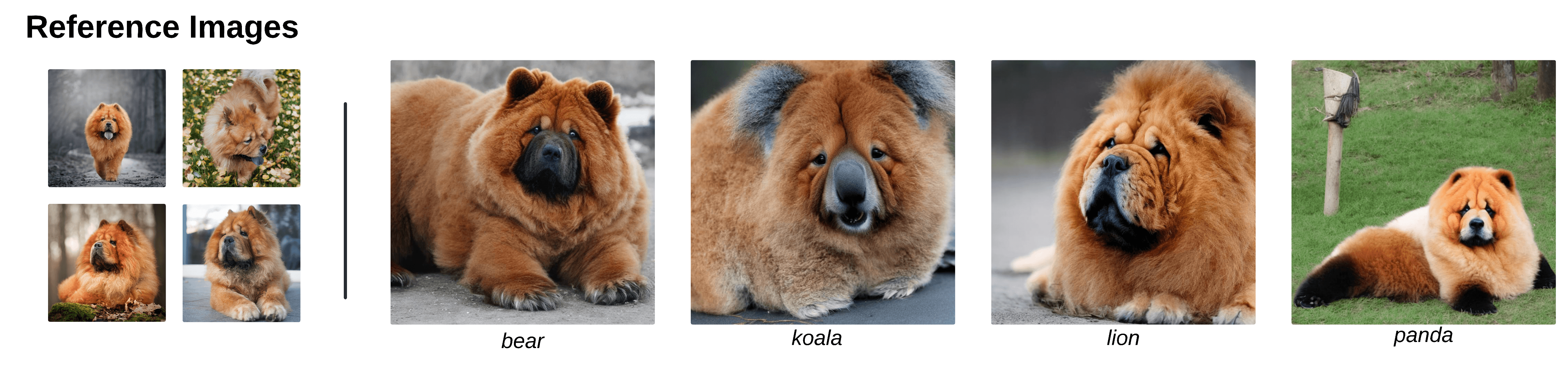}
    \caption{Novel hybrid synthesis samples from the RPO algorithm. We apply the prompts \texttt{``a cross of a [V] chow chow and [target species]''} to the RPO-trained model to generate these images. We highlight that RPO can synthesize new hybrids that retain the identity of the subject chow chow and perform property modifications.}
    \label{fig:novel_hybrids}
\end{figure}

\begin{figure}
    \centering
    \includegraphics[width=\textwidth]{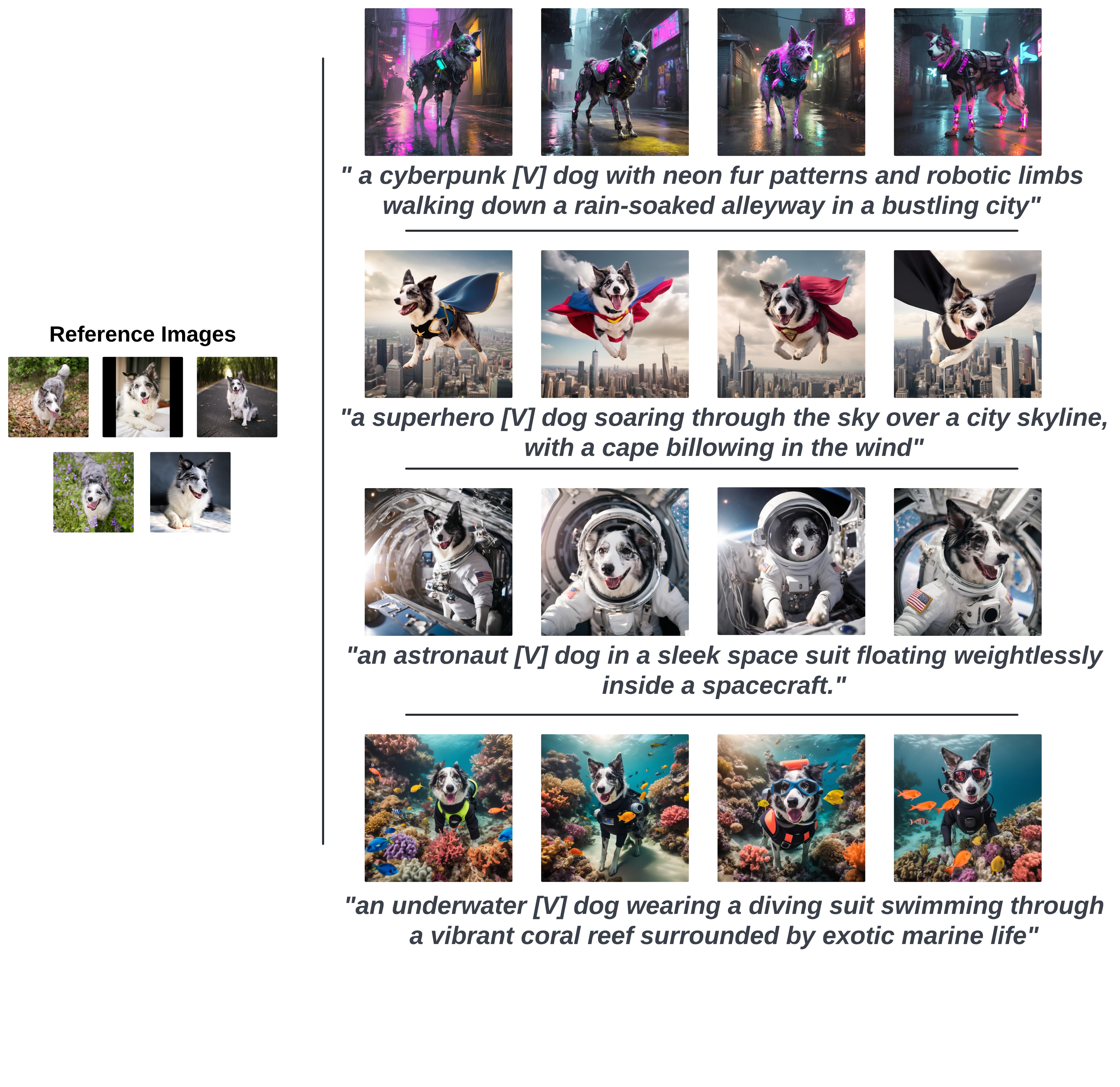}
    \caption{Novel prompts results from the RPO-trained model. We use ChatGPT to create 4 different highly imaginative prompts for a specific dog. We select 4 images for each prompt. The results demonstrate the generalization capability of RPO for handling novel prompts.}
    \label{fig:novel_prompts_dog}
\end{figure}

The skill set of the RPO-trained model is varied and includes re-contextualization (Figure~\ref{fig:recontext}), artistic style generation (Figure~\ref{fig:art_renditions}), expression modification (Figure~\ref{fig:expression_modification}), subject accessorization (Figure~\ref{fig:outfig_accessories}), color editing (Figure~\ref{fig:color_editing}), multi-view rendering (Figure~\ref{fig:multiple_view}), novel hybrid synthesis (Figure~\ref{fig:novel_hybrids}), and novel prompt generation (Figure~\ref{fig:novel_prompts_dog}).

\end{document}